\RequirePackage{fix-cm}
\documentclass[smallcondensed]{svjour3}     
\smartqed  
\usepackage{graphicx}
\usepackage{verbatim}
\usepackage{afterpage}
\usepackage{url}
\usepackage{amsmath}
\usepackage{enumerate}
\usepackage[round]{natbib}
\usepackage{inputenc}
\usepackage{sidecap}
\usepackage{booktabs}
\usepackage{todonotes}
\usepackage{xfrac}
\usepackage{placeins}

\journalname{Swarm Intelligence}
\begin{document}
\sloppy

\title{Evolution of Swarm Robotics Systems with \\Novelty Search}

\author{Jorge Gomes \and Paulo Urbano \and \\Anders Lyhne Christensen}

\institute{J. Gomes \at
              LabMAg, FCUL \& Instituto de Telecomunica\c{c}\~oes, Lisbon, Portugal \\
              \email{jgomes@di.fc.ul.pt}
           \and
           P. Urbano \at
           	  LabMAg, FCUL, Lisbon, Portugal \\
              \email{pub@di.fc.ul.pt}
           \and
           A. L. Christensen \at
              Instituto Universit\'ario de Lisboa (ISCTE) \& Instituto de Telecomunica\c{c}\~oes, Lisbon, Portugal\\
           	  \email{anders.christensen@iscte.pt}
}

\date{Received: 2012 / Accepted: April 2013}

\maketitle

\begin{abstract}

Novelty search is a recent artificial evolution technique that challenges traditional evolutionary approaches. In novelty search, solutions are rewarded based on their novelty, rather than their quality with respect to a predefined objective. The lack of a predefined objective precludes premature convergence caused by a deceptive fitness function. In this paper, we apply novelty search combined with NEAT to the evolution of neural controllers for homogeneous swarms of robots. Our empirical study is conducted in simulation, and we use a common swarm robotics task -- aggregation, and a more challenging task -- sharing of an energy recharging station. Our results show that novelty search is unaffected by deception, is notably effective in bootstrapping the evolution, can find solutions with lower complexity than fitness-based evolution, and can find a broad diversity of solutions for the same task. Even in non-deceptive setups, novelty search achieves solution qualities similar to those obtained in traditional 
fitness-based evolution. Our study also encompasses variants of novelty search that work in concert with fitness-based evolution to combine the exploratory character of novelty search with the exploitatory character of objective-based evolution. We show that these variants can further improve the performance of novelty search. Overall, our study shows that novelty search is a promising alternative for the evolution of controllers for robotic swarms.

\end{abstract}


\section{Introduction}


Motivations for the use of evolutionary techniques to design control systems for robots are numerous~\citep{harvey93,nelson09}. In the swarm robotics domain in particular, the complexity stemming from the intricate dynamics required to produce self-organised behaviour complicates the hand-design of control systems. Artificial evolution, on the contrary, has been shown capable of exploiting the intricate dynamics and synthesise self-organised behaviours~(see for example \citealp{trianni06b,sperati08}). However, evolutionary robotics techniques have only been proven effective in relatively simple tasks~\citep{sprong11,doncieux11,brambilla12}. The lack of studies that report successful evolution of behaviours for complex tasks can be ascribed to the difficulties in configuring the evolutionary process such that adequate solutions are synthesised within a reasonable amount of time \citep{doncieux11}.

The most common approach in artificial evolution, and evolutionary robotics in particular, is to guide the evolutionary process towards a fixed objective \citep{nelson09} (henceforth referred to as \emph{fitness-based evolution}). The experimenter defines a fitness function that estimates the quality of candidate solutions with respect to a given task, and this fitness function is used to score the individuals in the population. While search based on the task objective may intuitively seem reasonable, it is associated with a number of issues of which \emph{deception} is one of the most prominent~\citep{whitley91,jones95}. Deception is a challenging issue in evolutionary computation which occurs when the fitness function misguides the evolutionary process \citep{stan11a}, potentially causing evolution to converge to local optima. As the complexity of a task or a system increases, it becomes more difficult to craft an appropriate fitness function, and fitness-based evolution becomes more vulnerable to deception \citep{zaera96}.

Novelty search \citep{stan11a} is a distinctive evolutionary approach which rewards solutions based solely on their behavioural novelty. In fitness-based evolution, the objective is typically static, and the evaluations of individuals are independent of one another. In novelty search, on the other hand, individuals are evaluated by a dynamic measure that scores candidate solutions based on how different they are from solutions evaluated so far, with respect to their behaviour. Due to the absence of a static objective, novelty search is unaffected by premature convergence. \citet{mouret09} showed that the novelty-based paradigm can also be effective in bootstrapping evolution. Novelty search has proven capable of finding a broad diversity of solutions to a given problem \citep{stan11b} and solutions with lower neural network complexity than fitness-based evolution \citep{stan11a}. Novelty search has been successfully applied to many domains, including non-collective evolutionary robotics~(for examples, see \citealp{krcah10,stan11a,mouret11}).

In this paper, we propose and study the application of novelty search to the evolution of neural controllers for swarm robotics systems. Our motivation is the high level of complexity associated with swarm robotics, which stems from the intricate dynamics between many interacting units. The high level of complexity has the tendency to generate deceptive fitness landscapes~\citep{whitley91}, and novelty search has been shown to be unaffected by deception~\citep{stan11a}. To evaluate novelty search in the swarm robotics domain, we conduct several experiments. We use two different tasks in our study: (i)~an aggregation task, and (ii)~a resource sharing task. The former is a task commonly used in the field of evolutionary swarm robotics. The latter is a more challenging task in which the swarm must coordinate to ensure that each member has periodic and exclusive access to a charging station. In all experiments, we establish comparisons between novelty search and traditional fitness-based evolution. One of the key components in novelty search is the \emph{novelty measure} that quantifies the novelty of each solution. It is based on a \emph{behaviour characterisation} (usually a real-valued vector) that corresponds to an approximate representation of the individual's actual behaviour. The behaviour characterisation is typically domain-dependent and task-dependent. We study different approaches to the definition of behaviour characterisations for swarm robotics tasks. Our characterisations capture the macroscopic swarm-level behaviour and thus are independent of the swarm size. In the aggregation task, we evaluate characterisations composed of behavioural features sampled at regular intervals. In the resource sharing task, we go on to show that simple characterisations that summarise an entire simulation are, in fact, sufficient for novelty search to find good solutions.

One issue that can arise in novelty search is that a significant part of the effort may be spent exploring novel but unfruitful regions of the behaviour space \citep{stan10a,cuccu11b}. A number of methods that combine the exploratory nature of novelty search with the exploitatory nature of fitness-based evolution have been proposed to address this issue \citep{stan10a,cuccu11b,mouret11,gomes12}. We explore the potential of two such methods, namely \emph{progressive minimal criteria novelty search}~\citep{gomes12} and \emph{linear scalarization} of novelty and fitness objectives~\citep{cuccu11b}. 

The paper is organised as follows: in Section~\ref{sec:related_work}, we discuss related work, present the current challenges in evolutionary robotics, and introduce the novelty search algorithm. In Section~\ref{sec:aggregation}, we use novelty search to evolve aggregation behaviours. We experiment with three behaviour characterisations, and study how each one affects the behavioural diversity and the performance of novelty search. In Section~\ref{sec:sharing}, we experiment with a more challenging resource sharing task. We find that some behaviour characterisations open the search space too much and thereby reduce the effectiveness of novelty search. In Section~\ref{sec:combination}, we show how the problem of vast behaviour spaces can be mitigated by combining novelty search with fitness-based evolution. We conclude in Section~\ref{sec:conclusion} with a summary of the contributions of the paper and with a discussion of ongoing work.

\section{Related Work}
\label{sec:related_work}

In this section, we first discuss swarm robotics and evolutionary robotics, and the main challenges associated with these fields. We then present novelty search and how it can overcome some of these challenges. We go on to review recently proposed variants of novelty search. We conclude the section with a description of NEAT, the neuroevolution method used in our experiments.

\subsection{Swarm Robotics}

The field of swarm robotics, as well as the more general field of swarm intelligence~\citep{BonDorThe99:book}, take inspiration from the observation of social insects. In a swarm intelligence system, be it natural such as an ant colony, or artificial such as a large-scale decentralised multirobot system, relatively simple units rely on self-organisation to display collectively intelligent behaviour. As such, swarm robotics is an auspicious approach to the decentralised coordination of large numbers of robots~\citep{sahin05a}. An extensive survey of the modelling of swarm robotics systems and the problems that have been addressed can be found in \citep{brambilla12,sahin07}. Self-organisation in multirobot systems has, however, proven difficult to design by hand. Manually designing the control for the individual units of a swarm requires the decomposition of the macroscopic swarm behaviour into microscopic behavioural rules \citep{trianni06b}. Such decomposition includes discovering the relevant interactions between the individual robots, and between the robots and the environment, which will ultimately lead to the emergence of global self-organised behaviour. Unfortunately, there is no general method for decomposing a desired global behaviour into the rules that govern each individual. System designers therefore typically take inspiration from biological swarm systems or rely on manual trial and error.

\subsection{Evolutionary Robotics}

Evolutionary robotics is a field concerned with the application of evolutionary computation to the synthesis of robotic systems. Evolutionary robotics is an alternative for the design of control for swarm robotics systems, because the application of evolutionary computation eliminates the need for manual decomposition of the desired macroscopic behaviour. Artificial evolution essentially performs an iterative trial and error process in which candidate solutions are evaluated according to their swarm-level behaviour. Macroscopic performance evaluation is thus used to guide the evolutionary process towards the objective. Several swarm robotics tasks have been solved with evolutionary approaches, such as coordinated motion~\citep{baldassarre07}, foraging~\citep{liu07}, aggregation~\citep{trianni03}, hole avoidance~\citep{trianni06a}, aerial vehicles communication \citep{hauert09}, categorisation~\citep{ampatzis08}, group transport~\citep{gross08}, and social learning \citep{pini08}. 

Traditional evolutionary approaches are, however, prone to suffer from a number of issues \citep{doncieux11}. Deception~\citep{whitley91,jones95} is a challenging issue in evolutionary computation, because it can cause the evolutionary process to converge prematurely to local optima. Deception occurs when the fitness function creates a \emph{deceiving} fitness gradient. This typically happens when the fitness function fails to adequately reward the intermediated steps that are needed to achieve the global optimum. A related issue that can arise when applying evolutionary computation to complex tasks is the bootstrap problem \citep{gomez96,mouret09}. This problem occurs when the task is too demanding to exert significant selective pressure on the population during the early stages of evolution, as all of the individuals perform equally poorly. As a consequence, there is no fitness gradient and the evolutionary process starts to drift in an uninteresting region of the solution space.

One way to circumvent deception is through the use of techniques that maintain genotypic diversity in the population, such as fitness sharing~\citep{goldberg87b}, promotion of diversity based on the fitness score of the solutions~\citep{hu05,hutter06}, intermingling individuals of different genetic ages~\citep{hornby06,castelli11}, and minimisation of the age of the genotypes \citep{schmidt10}. However, the problem may ultimately be in the fitness function itself, not in the particular search algorithm. If the fitness function is actively misguiding the search, the evolutionary process may still fail, regardless of the amount of genotypic diversity present in the population. 

A distinct approach to address the issue of deception is through the use of coevolution. In competitive coevolution, individual fitness is evaluated through competition with other individuals in the population, rather than through an absolute fitness measure. Ideally, this creates an \emph{arms race} that leads to increasingly better solutions. This approach has been applied with success in some domains (e.g. \citet{chellapilla99}), but it is also associated with a number of issues that stem from the potentially counterproductive dynamics between the multiple co-evolving species, such as convergence to mediocre stable states \citep{watson01}. Other techniques to overcome deception and to bootstrap evolution rely on the decomposition of the objective into multiple sub-goals that each are easier to attain. These techniques include incremental evolution~\citep{gomez96}, fitness shaping~\citep{uchibe02}, and multi-objectivisation~\citep{deb01,knowles01}. A common drawback of these approaches is that task decomposition may not always be possible, and when it is, a significant amount of \emph{a priori} knowledge about the task is required to devise appropriate sub-tasks.

\subsection{Novelty Search}
\label{sec:ns}

While the methods discussed above for mitigating deception might help the evolutionary process to avoid getting stuck in local optima, they leave the underlying problem untreated, namely that the fitness function itself might be misdirecting the search. With this issue in mind, a new evolutionary approach was recently proposed --- novelty search. In novelty search, the evolutionary process is based on the promotion of phenotypic (i.e., behavioural) diversity and innovation, contrasting with more common techniques that strive to maintain genotypic diversity. \citet{stan11a} used two deceptive robotics tasks to show how novelty search was able to find good solutions faster and more consistently than fitness-based evolution. Even though the objective was not directly pursued in any of their experiments, a solution was found more consistently through the exploration of the behaviour space. Successful applications of novelty search include the evolution of adaptive neural networks \citep{soltoggio09}; genetic programming \citep{stan10b}; evolution strategies \citep{cuccu11a}; body-brain co-evolution \citep{krcah10}; biped robot control \citep{stan11a}; and robot navigation in deceptive mazes \citep{stan08, mouret11}.

Implementing novelty search requires little change to any evolutionary algorithm aside from replacing the fitness function with a domain-dependent novelty metric. This metric quantifies how different an individual is from the other, previously evaluated individuals with respect to behaviour. Previously seen behaviours are stored in an archive. The archive is initially empty, and new behaviours are added to it if they are significantly different from the ones already there, i.e., if their novelty score is above some threshold.

A novelty metric characterises how far an individual is from other individuals in the behaviour space. This metric depends on the sparseness at a given point in the behaviour space. A simple measure of sparseness at a point is the average distance to the $k$-nearest neighbours of that point, where $k$ is a fixed parameter empirically determined. The sparseness $\rho$ at point $x$ is given by:
\begin{equation}
\rho (x)=\frac{1}{k}\sum_{i=1}^{k}dist(x,\mu _{i}) \enspace ,
\end{equation}
where $\mu _{i}$ is the $i$th-nearest neighbour of $x$ with respect to the distance metric $dist$. Note that the computational cost of the nearest-neighbours calculation increases linearly with the size of the population and the size of the archive. However, it is possible to limit the size of the archive \citep{stan11a} and to use data structures such as KD-trees to reduce this cost. The function $dist$ is a measure of behavioural difference between two individuals in the search space. Candidates from sparse regions of the behaviour space thus tend to receive higher novelty scores, which results in an evolutionary process that strives to uniformly explore the behavioural space. Note that the novelty metric promotes behavioural diversity within the population at all times and therefore helps to avoid convergence to a single solution, which is common in fitness-based evolution.

The behaviour of each individual is typically characterised by a vector of real numbers. The behavioural distance $dist$ is then given by the distance between the corresponding characterisation vectors (the Euclidean distance is typically used). The experimenter should design the behaviour characterisation so that it captures behaviour aspects that are considered relevant to the problem or task. For example, in a maze navigation task \citep{stan11a} the behaviour characterisation was the trajectory of the robot through the maze. The design of the characterisation has direct implications on the effectiveness of novelty search. An excessively detailed characterisation can open the search space too much, and might cause the evolution to focus on regions of the behaviour space that are irrelevant for the task for which a solution is sought. On the other hand, an incomplete or inadequate characterisation can originate counterproductive \emph{conflation}. Conflation occurs because the mapping between observable behaviours and behaviour characterisations is typically not injective. As such, notably different behaviours can have similar behaviour characterisations, which can potentially hinder the evolution of novel solutions \citep{kistemaker11}.

Designing the behaviour characterisation often requires knowledge about which behaviour features are relevant for solving a task. However, unlike fitness shaping techniques, it is not necessary to understand exactly how these behaviour features affect fitness, or in which order the features must be evolved. Novelty search does not require a fitness gradient to guide evolution, which makes the approach applicable to some classes of problems that are difficult to solve using traditional fitness-based evolution \citep{kistemaker11}.

\subsection{Novelty Search Variants} \label{sec:ns_variants}

Several extensions of novelty search have been proposed to overcome the limitation of novelty search with respect to guiding evolution towards good solutions in vast behaviour spaces. These extensions are based on the combination of the exploratory character of novelty search with the exploitatory character of fitness-based evolution.

\citet{stan10a} proposed \emph{minimal criteria novelty search}~(MCNS), an extension of novelty search where individuals must meet some domain-dependent minimal criteria to be selected for reproduction. In \citep{stan10a}, the authors applied MCNS in two maze navigation tasks and demonstrated that MCNS evolved solutions more consistently than both novelty search and fitness-based evolution. Similar results were reported in \citep{kirkpatrick12}, using competitive coevolution and a different task. However, MCNS suffers from a number of drawbacks \citep{stan10a}. First, the choice of minimal criteria in a particular domain requires careful consideration and domain knowledge, since it adds significant restrictions to the search space. Constraining the search space too much can hinder the evolution of some types of solutions. Second, if no individuals are found that meet the minimal criteria, search is effectively random. In situations where an initial randomly generated population is unlikely to contain such individuals, it may therefore be necessary to seed MCNS with a genome specifically evolved to meet the criteria. And finally, if the minimal criteria are too stringent, it might be difficult to mutate apt individuals without violating the criteria, and thus many evaluations may be wasted.

In recent work \citep{gomes12}, we proposed an extension named \emph{progressive minimal criteria novelty search} (PMCNS), which overcomes the drawbacks of MCNS. In PMCNS, the respective benefits of novelty search and fitness-based evolution are combined by letting novelty search freely explore new regions of the behaviour space, as long as solutions meet a progressively stricter fitness criterion. PMCNS was found to outperform several other evolutionary algorithms, and to evolve higher scoring individuals while still maintaining behavioural diversity.

\citet{cuccu11b} proposed an alternative approach for combining novelty and fitness, where the score of each individual is based on a linear scalarization of the novelty score and the fitness score (henceforth referred to as \emph{linear scalarization}). They applied the approach to a deceptive box-pushing task, and found that linear scalarization outperformed both novelty search and fitness-based evolution. \citet{mouret11} proposed novelty-based multiobjectivisation, which is a Pareto-based multi-objective evolutionary algorithm. A novelty objective is added to the task objective in a multi-objective optimisation. The technique was applied to a deceptive maze navigation problem. Compared with pure novelty search, the use of multiobjectivisation only led to marginally better results. Other techniques for sustaining behavioural diversity in evolutionary robotics are reviewed in \citep{mouret12}.

\subsection{NEAT}
In our experiments, the controllers of the robots are time recurrent neural networks evolved by NEAT (short for NeuroEvolution of Augmenting Topologies) \citep{stan02}. NEAT is a widely used neuroevolution approach, and one of the most successful approaches developed to date. NEAT simultaneously optimises the weighting parameters and the structure of artificial neural networks. It begins the evolution with a population of small, simple networks and complexifies the network topology into diverse species over generations. This leads to the evolution of increasingly sophisticated behaviour. A key feature in NEAT is its distinctive approach to maintain a diversity of growing structures simultaneously. Unique historical markings are assigned to each new structural component. During crossover, genes with the same historical markings are aligned, producing valid offspring efficiently, without the need of complex topological comparisons. NEAT uses speciation and fitness sharing to protect new structural innovations. This reduces competition between networks with distinct topologies, providing time for the weights of new structures to be optimised. Networks are assigned to species based on the extent to which they share historical markings. Complexification is thus supported by both historical markings and speciation, allowing NEAT to establish high-level features early in evolution and then elaborate on them as the evolutionary process progresses. In effect, NEAT searches for a compact, appropriate network topology by incrementally complexifying existing structures.

It is important to note that both novelty search and NEAT strive to maintain diversity, but at different levels: whereas NEAT maintains \emph{genotypic} diversity, novelty search maintains \emph{phenotypic} diversity. Novelty search and NEAT thus complement one another and their combined use bears a number of advantages. In particular, the complexification mechanism of NEAT can introduce order in the exploration done by novelty search, with less complex behaviours being explored before progressing to more complex ones~\citep{stan11a}.

\section{Evolution of Aggregation Behaviours with Novelty Search}
\label{sec:aggregation}

In this section, we apply novelty search to the problem of swarm aggregation --- a commonly studied task in swarm robotics. In the aggregation task, a dispersed robot swarm must form a single cluster. We conduct three sets of experiments with novelty search, each with a distinct behaviour characterisation. We compare the performance of novelty search to the performance of traditional fitness-based evolution. We also include results from evolutionary runs with NEAT in which random fitness scores are assigned to individuals. Random evolution serves as a baseline for performance comparisons.

Aggregation is a task of fundamental importance in many biological systems. It is the basis for the emergence of various forms of cooperation, and can be a considered a prerequisite for the accomplishment of many collective tasks~\citep{trianni03}. Several works describe the evolution of aggregation behaviours for swarms of robots. Commonly, the parameters of neural networks with fixed topologies are optimised by fitness-based evolutionary algorithms. \citet{baldassarre03} successfully evolved controllers for a swarm of robots to aggregate and move towards a light source in a clustered formation. Three classes of behaviours were evolved, but each evolutionary run always converged to only one of the classes. \citet{trianni03} studied the evolution of a swarm of simple robots to perform aggregation in a square arena. Two different behaviours were evolved: \emph{static clustering} which leads to the formation of compact and stable clusters, and \emph{dynamic clustering} which leads to loose but moving clusters. \citet{sahin05b} used a similar experimental setup as~\citep{trianni03}, and studied how some of the parameters of the evolutionary algorithm affected the performance and the scalability of the evolved behaviours.

In the previous studies discussed above, the robots used directional sound sensing. Directional sensing allowed the robots to follow gradients towards groups of other robots emitting sound signals. The use of sound and directional microphones makes the aggregation task sufficiently easy for controllers based on reactive neural networks without any hidden neurons to solve the task~\citep{trianni03, baldassarre03, sahin05b}. In our work, the aggregation task is more challenging: the robots do not use sound, the range of the sensors is significantly lower than in previous studies, and the arena is larger. These modifications increase the difficulty of the task and may require radically different strategies for aggregation~\citep{soysal07}, since robots can only sense one another when they are close.

\subsection{Experimental Setup}
\label{sec:aggregation_setup}

Our experimental framework is based on the Simbad 3d Robot Simulator~\citep{hugues06} for the robotic simulations, and on NEAT4J\footnote{NeuroEvolution for Augmenting Topologies for Java -- \url{http://neat4j.sourceforge.net}} for the implementation of NEAT. Simbad 3D simulates kinematics and implements simple collision handling. The environment is a 3\,m by 3\,m square arena bounded by walls. The swarm is homogeneous and composed of 7 robots. The robots are modelled based on the e-puck educational robot~\citep{mondada09}, but do not strictly follow its specification. Each robot is circular with a diameter of 8\,cm and can move at speeds of up to 12\,cm/s. Regarding sensors, each robot is equipped with 8 IR sensors evenly distributed around its chassis for the detection of obstacles (walls or other robots) within a range of 10\,cm, and 8 sensors dedicated to the detection of other robots within a range of 25\,cm. Both types of sensors return the distance of the object that is being sensed, or the maximum value if nothing is sensed. An additional sensor (count sensor) returns the percentage of nearby robots (within a radius of 25\,cm), relative to the desired cluster size (the total swarm size in our experiments). The simulated sensors are not based on any specific hardware. Nevertheless, the obstacle sensors could be implemented with active IR sensors, while the robot sensors and the robot count sensor could be implemented with short-range communication~\citep{correll07,gutierrez2008}. The inputs of the neural network controller are the normalised readings from the sensors mentioned above. The controller has three outputs: one to control the speed of each motor, and one dedicated to completely stopping the robot if its activation is above 0.5.

We evaluate each controller 10 times. In each simulation, we vary the initial position and orientation of each robot. The initial positions are randomised in such a way that the robots are placed at least 50\,cm from one another, which ensures that the robots always are reasonably well distributed at the beginning of the simulation. Each simulation lasts for 2500 simulation steps, which corresponds to 250\,s of simulated time. The best individual of each generation was post-evaluated in 100 simulations, in order to obtain a more accurate fitness estimate.

\subsection{Configuration of the Evolutionary Algorithms}
\label{sec:aggregation_algsetup}

Fitness-based evolution and random evolution use the default NEAT implementation provided by the NEAT4J library. In random evolution, random fitness scores are assigned to each individual. Novelty search was implemented over NEAT, following the description and parameters in~\citep{stan11a}. We used a $k$ value of 15 nearest neighbours and individuals are stochastically added to the archive with a probability of 2\%, as suggested in \citep{stan10b}. The parameters for NEAT were the same in all experiments: recurrent links are allowed, crossover rate -- 25\%, mutation rate -- 10\%, population size -- 200, and each evolutionary process was conducted for 250 generations. The remaining parameters were assigned their default value according to the NEAT4J implementation.

The fitness function is based on the average distance to the centre of mass~(also used in \citep{trianni03}). The fitness $F_{a}$ of a simulation with $T$ time steps and $N$ robots is defined as:
\begin{equation}
F_{a}=\sum_{i=1}^{N}\frac{1-dist(\mathbf{R}_T,\mathbf{r}_{i,T})}{N} \enspace ,
\end{equation}
where $\mathbf{R}_T$ is the centre of mass at the end of the simulation, and $\mathbf{r}_{i,T}$ is the position of robot $i$ at the same instant. The distance values are normalised to $[0,1]$. The fitness scores obtained in each of the 10 simulations are combined to a single value using the harmonic mean, as advocated in~\citep{sahin05b}.

The behaviour characterisations we use in novelty search are based on spatial inter-robot relationships, measured at regular intervals of 5\,s throughout the simulation. We devised three characterisations:

\begin{description}
\item[$\mathbf{b_{cm}}$:] The average distance to centre of mass of the swarm is sampled throughout the simulation. Considering a simulation with $N$ robots and $\tau$ temporal samples, the behaviour characterisation $\mathbf{b_{cm}}$ is given by:
\begin{equation}
\mathbf{b_{cm}}=\frac{1}{N} \left [ \sum_{i=1}^{N}dist(\mathbf{R}_1,\mathbf{r}_{i,1}),\cdots , \sum_{i=1}^{N}dist(\mathbf{R}_\tau,\mathbf{r}_{i,\tau}) \right ] \enspace .
\label{eq:cm}
\end{equation}

\item[$\mathbf{b_{cl}}$:] The number of robot clusters is sampled at regular intervals throughout the simulation, inspired by the metric used in \citep{sahin05b}. Two robots belong to the same cluster if the distance between them is less than their sensor range (25\,cm). The behaviour characterisation $\mathbf{b_{cl}}$ is given by:
\begin{equation}
\mathbf{b_{cl}}=\frac{1}{N}\left [clustersCount(1), \cdots , clustersCount(\tau) \right ] \enspace .
\label{eq:cl}
\end{equation}

\item[$\mathbf{b_{cmcl}}$:] The two characterisations $\mathbf{b_{cm}}$ and $\mathbf{b_{cl}}$ are concatenated to form a single characterisation. Both $\mathbf{b_{cm}}$ and $\mathbf{b_{cl}}$ have the same length and each element of the characterisation vectors ranges from 0 to 1. Thus both components of the behaviour characterisation approximately have the same contribution to the novelty metric. The new characterisation $\mathbf{b_{cmcl}}$ is given by:
\begin{equation}
\mathbf{b_{cmcl}}=(\mathbf{b_{cm}},\mathbf{b_{cl}}) \enspace .
\label{eq:comb}
\end{equation}
\end{description}

We computed the spatial inter-robot relationships at every 5\,s and the simulation lasted for 250\,s. This resulted in behaviour characterisation vectors of length 50 for $\mathbf{b_{cm}}$ and $\mathbf{b_{cl}}$, and vectors of length 100 for $\mathbf{b_{cmcl}}$. As 10 simulations are conducted to evaluate each controller, the corresponding final behaviour characterisation vector is the element-wise mean of the vectors obtained in all 10 simulations. In order to establish a basis for comparison, all the controllers evolved by novelty search and by random evolution were also scored by the fitness function $F_{a}$. It is important to note that the fitness scores did not have any influence on the evolutionary process in the novelty search experiments.

\subsection{Performance Comparison}

\begin{figure}
	\centering
	\includegraphics[width=1\textwidth]{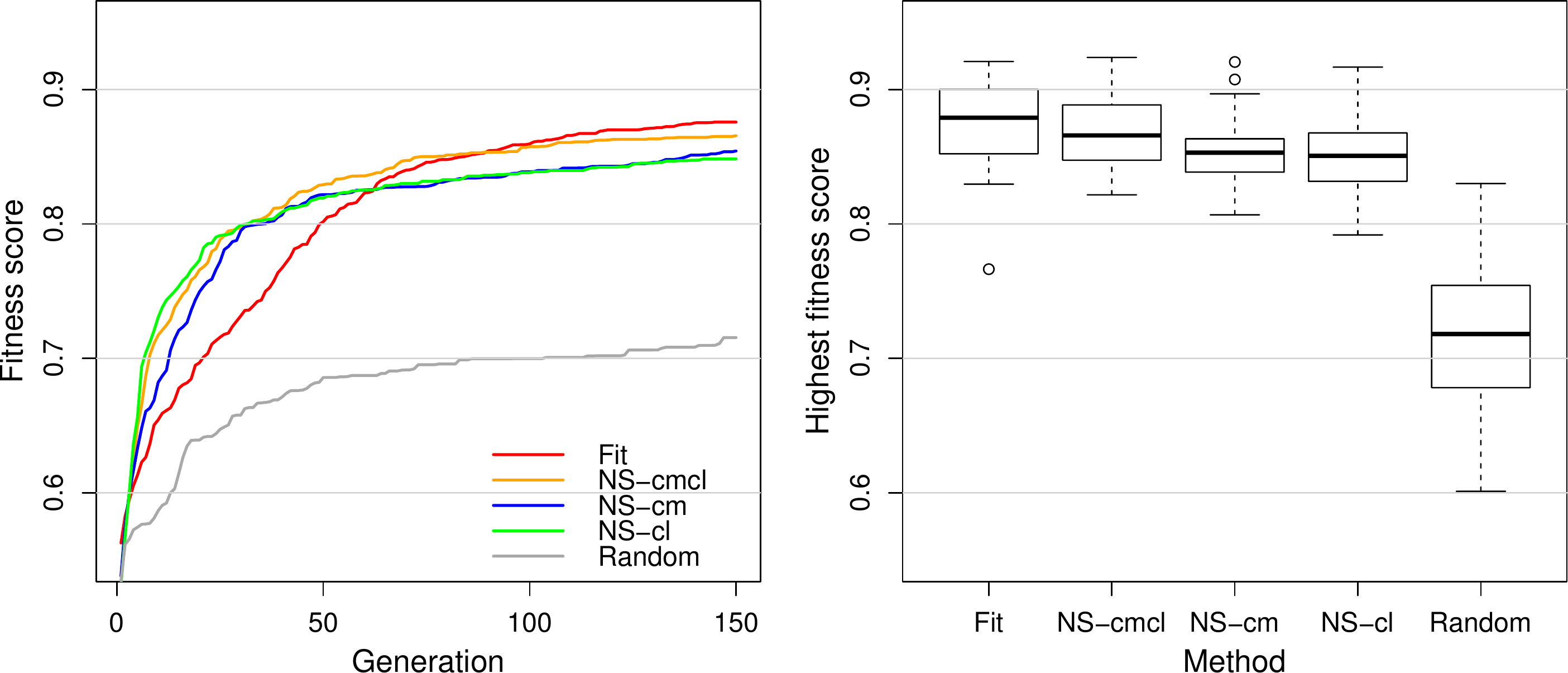}
\caption{Highest fitness scores achieved in the aggregation task with fitness-based evolution (\emph{Fit}), random evolution (\emph{Random}), and novelty search with the three behaviour characterisations (\emph{NS-cmcl}, \emph{NS-cm}, \emph{NS-cl}). Left: average fitness value of the highest scoring individual found so far at each generation. The values are averaged over 30 independent evolutionary runs for each method. Right: box-plots of the highest fitness score found in each evolutionary run, for each method. The whiskers extend to the lowest and the highest data point within 1.5 times the interquartile range. Outliers are indicated by circles.}
\label{fig:ccm_graph}
\end{figure}

The fitness trajectories for novelty search, fitness-based evolution, and random evolution, are depicted in Figure~\ref{fig:ccm_graph}(left). The data points plotted are the averages of the highest fitness score found so far from the start of the evolutionary run and until the current generation.\footnote{The values that are plotted are fitness scores measured by the fitness function $F_{a}$, and not the actual scores used for selection in novelty search or in random evolution. Note that even random evolution has an ascending fitness trajectory because the data points plotted are the highest score achieved so far in the evolutionary process. The trajectory for random evolution can thus increase from time to time when an individual, by chance, scores higher than any previously evaluated individual.} Note that since novelty search does not follow a fitness gradient, the best solutions are not necessarily found in the last generation. As such, it is necessary to save the interesting solutions (for instance, the best one found so far) throughout the evolutionary process. We continued the evolutionary process beyond the 150th generation for all experiments, but there was no significant change in the fitness scores after that point. Although the lowest possible fitness score is $\approx 0.05$, the highest fitness of an initial random population is on average $\approx 0.55$. The relatively high fitness of an initial random population is explained by the fact that stochastically moving robots tend be significantly closer to one another than in the worst case scenario where robots are located in opposite corners of the arena.

Figure~\ref{fig:ccm_graph}(right) shows box-plots of the highest fitness score for each method in 30 evolutionary runs. The results show that novelty search could consistently find relatively high scoring solutions, with all the behaviour characterisations. There was no significant differences between the highest fitness scores achieved with fitness-based evolution and with novelty search with $\mathbf{b_{cmcl}}$ (Mann--Whitney U test, $p$-value $<$ 0.05). With $\mathbf{b_{cl}}$, the highest fitness scores were significantly lower than the highest scores found by novelty search with $\mathbf{b_{cmcl}}$ and fitness-based evolution. The capacity of novelty search to bootstrap the evolutionary process should be noted. From around generation 5 to generation 40, all novelty search variants achieve fitness scores significantly higher than fitness-based evolution ($p$-value $<$ 0.01).

Previous studies have shown that novelty search can perform better than fitness-based evolution in deceptive tasks, but fails to match the performance of fitness-based evolution when the task is non-deceptive \citep{stan11a,mouret11}. Our results reveal that in the aggregation task, the fitness-function is not deceptive, as fitness-based evolution typically converges to the most effective strategy, achieving high fitness scores (except in a single run, see Figure~\ref{fig:ccm_graph}). Still, novelty search managed to achieve fitness scores that are comparable to the scores achieved in fitness-based evolution. 

\subsection{Behavioural Diversity}

The analysis of the explored behaviour space in novelty search and in fitness-based evolution allows for a better understanding of the evolutionary dynamics. Since each behaviour characterisation vector has either 50 or 100 dimensions, we applied a dimensionality reduction method to facilitate visualisation of the explored regions of the behaviour space. We used a Kohonen self-organising map \citep{kohonen90}. Kohonen maps are neural networks trained using unsupervised learning to produce a two-dimensional discretisation of the input space of the training samples, while preserving the topological relations. We trained a Kohonen map with the behaviour vectors found both by novelty search and by fitness-based evolution, and then mapped each individual to the region whose vector is more similar to the individual's behaviour vector.

\subsubsection{Centre of mass behaviour characterisation} \label{sec:agg_cm}

\afterpage{
\begin{figure}
\centering 
	\includegraphics[width=1\textwidth]{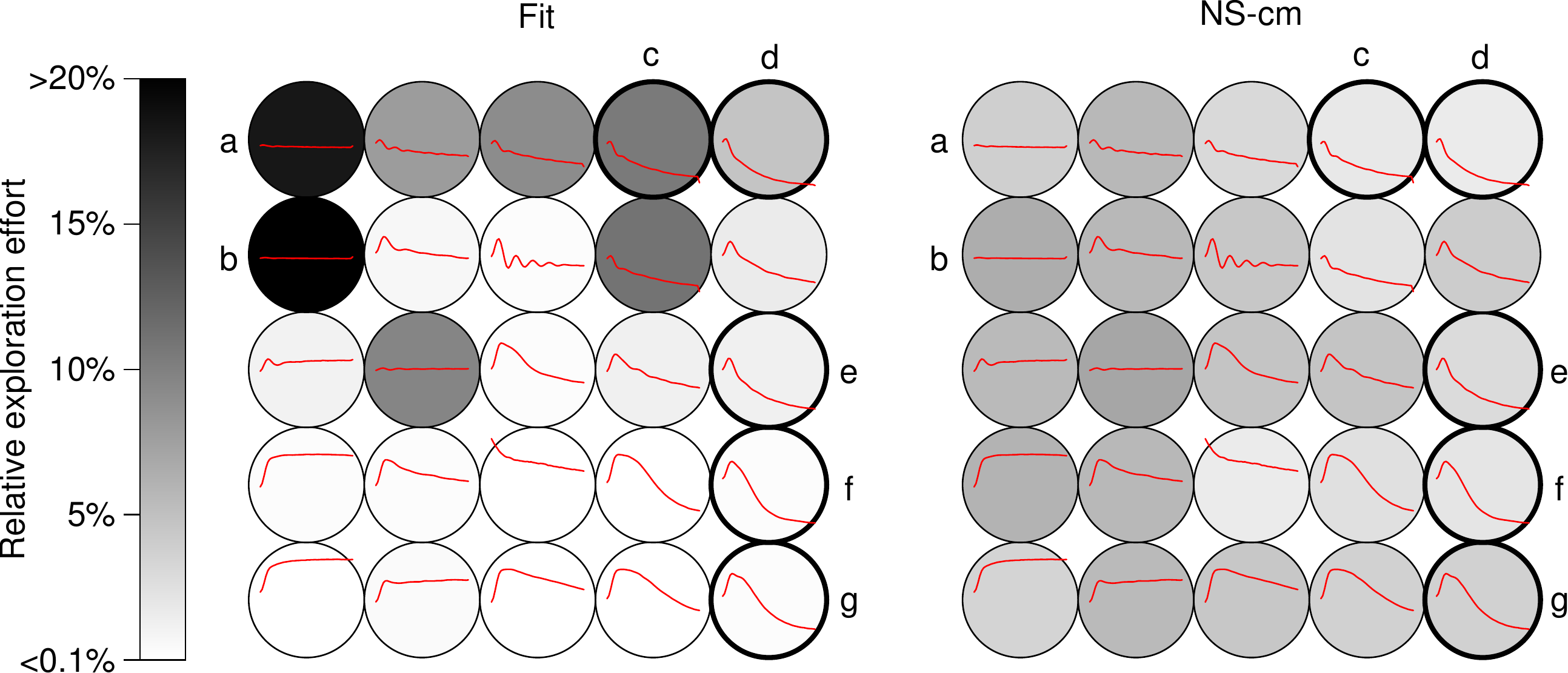}
\caption{Kohonen maps representing the explored behaviour space in fitness-based evolution (Fit) and in novelty search with $\mathbf{b_{cm}}$ (NS-cm). Each circle is a neuron corresponding to the vector depicted by the embedded plot (the average distance to the centre of mass over time). Each behaviour vector is mapped to the neuron with the most similar vector. The darker the background of a neuron is, the more behaviours were mapped to it. The regions associated with higher fitness scores are indicated with a bold circle (regions \emph{c} to \emph{g}).}
\label{fig:som_cm}
\end{figure}

\begin{table}
\caption{The number of individuals mapped to each behaviour region \emph{a-g} (see Figure~\ref{fig:som_cm}), relative to the total number of mapped individuals.}
\centering
\begin{tabular}{r r r}
\toprule
Behaviour region & Fit exploration (\%) & NS-cm exploration (\%) \\
\midrule
a & 18.17 & 3.80 \\
b & 22.18 & 6.39 \\
c & 10.27 & 1.82 \\
d & 4.57 & 1.54 \\
e & 1.23 & 2.75 \\
f & 0.15 & 1.90 \\
g & 0.14 & 3.50 \\
\bottomrule
\end{tabular}
\label{tab:som_cm}
\end{table}
}

The Kohonen maps corresponding to the explored behaviour space in fitness-based evolution and novelty search with $\mathbf{b_{cm}}$ can be seen in Figure~\ref{fig:som_cm}. The results show that fitness-based evolution focused the search on only a subset of the behaviour regions. In contrast, novelty search explored the behaviour space much more uniformly. If we consider the behaviour regions associated with higher fitness scores (regions \emph{c} to \emph{g}), important differences become apparent: fitness-based evolution avoids behaviour regions where the average distance to the centre of mass rises beyond the initial value, such as the regions \emph{f} and \emph{g} (see Table~\ref{tab:som_cm}). Instead, the evolutionary process is much more focused on behaviours that lead to a monotonic decrease in the average distance to the centre of mass. This bias is introduced by the fitness function, as it favours a low average distance to centre of mass. Evolving solutions where the average distance rises beyond the initial value might require going against the fitness gradient. As such, these types of solutions are impeded in the fitness-based evolutionary process. Novelty search, on the other hand, is not subject to a static evolutionary pressure, and can therefore explore and discover a wider range of solutions to the task.

The analysis of the behaviour patterns (see plots inside the neurons in Figure~\ref{fig:som_cm}) reveals an interesting point: the regions explored by novelty search are characterised by vectors that differ much less than uniformly sampled vectors of length 50 would. This happens because the elements of a given behaviour vector are inherently correlated. Robots cannot, for instance, travel instantly to any location in the environment, and the difference between consecutive samples of the average distance to the centre of mass is therefore limited by the speed of the robots. As such, the reachable behaviour regions constitute only an (often small) subset of the total behaviour space. This explains why novelty search can consistently find successful solutions to the task, even in a high-dimensional behaviour space.

The results also show that fitness-based evolution spends a considerable amount of time in behaviour regions where there are no aggregation dynamics at all (regions \emph{a} and \emph{b}). These regions correspond to the initial best solutions, where robots do not move or move randomly in the arena, in order to maintain an average distance to the centre of mass that is at least as low as the average distance at the start of the simulation. This class of behaviours constitutes a local maximum. Novelty search does not spend as much time in such behaviour regions (see Table~\ref{tab:som_cm}), as the search moves towards regions with novel behaviours. Since the initial average distance to the centre of mass is situated in the middle of the spectrum, the novel behaviours can be aggregation behaviours as well as dispersion behaviours. Typically, each novelty search evolutionary run explores both types of behaviour simultaneously, maintaining a healthy diversity in the population. This phenomenon can offer an explanation for the relatively good performance of novelty search in the earlier stages of evolution: novelty search does not get stuck in the early local maximum, and as such can explore and discover better solutions faster.

\begin{figure}
\centering 
	\includegraphics[width=1\textwidth]{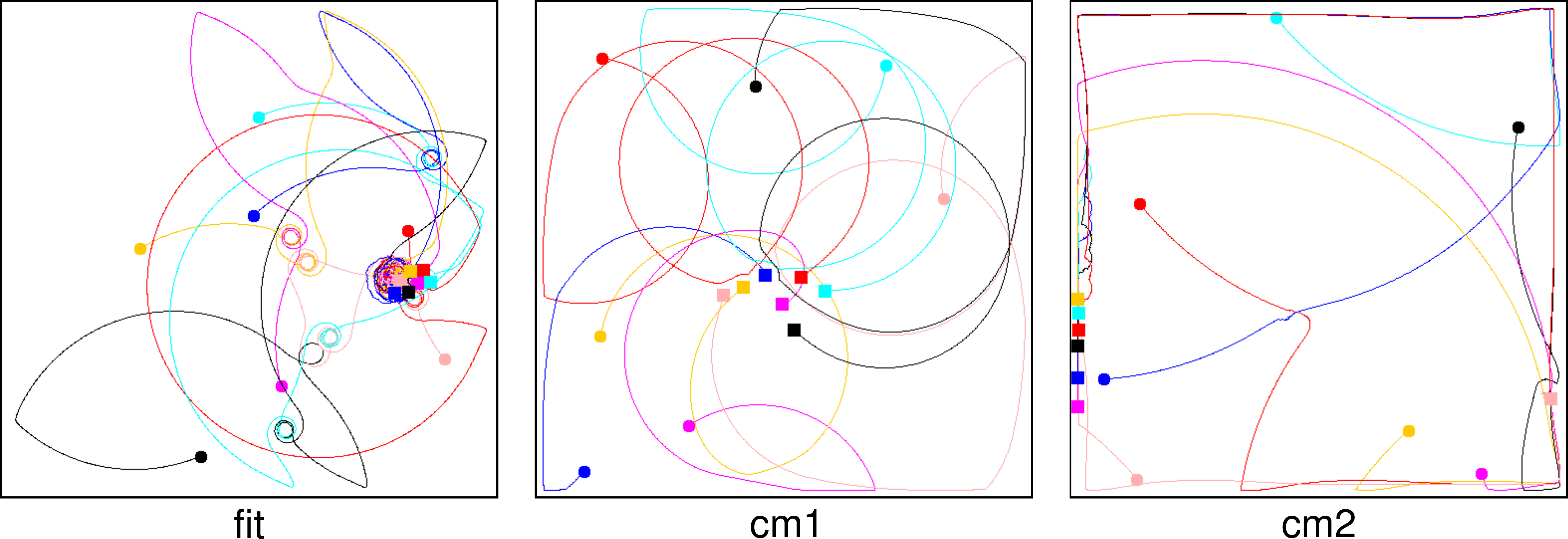}
\caption{The typical best solution evolved by fitness-based evolution (fit), and two examples of solutions found by novelty search with $\mathbf{b_{cm}}$ (cm1 and cm2). Each line represents the trajectory of a single robot throughout the simulation. The circles depict the initial positions of the robots and the squares depict their final positions. Videos of the behaviours are available as online supplemental material.}
\label{fig:behaviours_cm}
\end{figure}

To confirm the behavioural diversity evolved by each method, we resorted to the visual inspection of the best solutions. In fitness-based evolution, the highest scoring individuals tended to follow the same behaviour pattern (Figure~\ref{fig:behaviours_cm},~\textbf{fit}):

\begin{description}
\item[\bf fit:] The robots explore the environment in large circles, and form static clusters when they encounter one another. If a cluster is small, the robots abandon it after a while and recommence the circular exploration.
\end{description}

The \textbf{fit} behaviour pattern was also often found by novelty search. However, novelty search commonly evolved a different type of solutions in which walls are exploited to achieve aggregation. Examples of behaviour patterns that use the walls are described below and depicted in Figure~\ref{fig:behaviours_cm}~(\textbf{cm1} and \textbf{cm2}).

\begin{description}
\item[\bf cm1:] The robots go towards the walls while avoiding any other robots encountered. When they reach a wall, they follow the wall for a while and then depart with a certain circular trajectory that causes them to pass through the centre of the arena. Eventually, the robots end up forming a loose cluster close to the centre.
\item[\bf cm2:] The robots move in straight lines until they encounter a wall, and then, depending on the approach angle, they either remain static for a while or start to follow the wall. When two or more robots encounter one another, they stop and form a cluster near the wall.
\end{description}

Visual inspection of the behaviours confirms that fitness-based evolution did not explore some classes of solutions, and  in particular, solutions where the robots navigate near the walls. If all the robots initially move towards one of the walls surrounding the arena, they will often end up far apart. Consequently, the centre of mass of the robots will often be close to the centre of the arena, far from the robots, which explains the initial high average distance to the centre of mass (regions \emph{e, f, g} in Figure~\ref{fig:som_cm}). Learning to navigate near the walls potentially requires the evolution of many solutions with very low fitness scores. Avoiding navigation close to walls, on the other hand, results in higher fitness scores, because robots will, by chance, be closer to one another, and therefore to the centre of mass. Given the opportunistic nature of fitness-based evolution, the stepping stone of first navigating along walls to later achieve aggregation, is thus unlikely to be found. In fact, we have been unable to find reports in the literature on the evolution of aggregation behaviours that exploit walls. Such behaviours have, however, proven successful in biological systems, such as in self-organised aggregation of cockroaches \citep{jeanson05}.

\subsubsection{Number of clusters behaviour characterisation}

\begin{figure}
\centering 
	\includegraphics[width=0.66\textwidth]{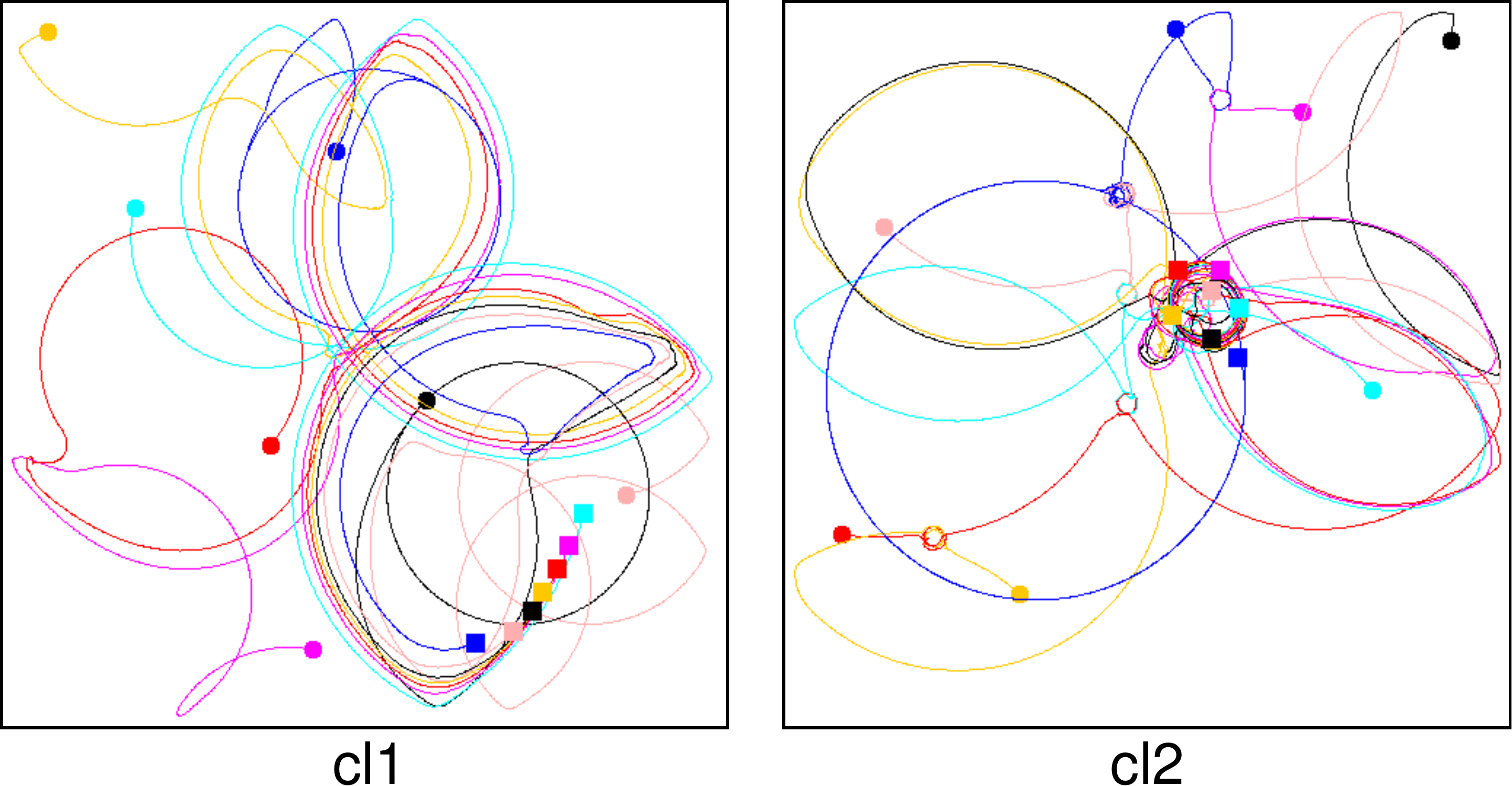}
\caption{Examples of distinctive behaviours evolved by novelty search with $\mathbf{b_{cl}}$. Each line represents the trajectory of a single robot throughout the simulation. The circles depict the initial positions of the robots and the squares depict their final positions. Videos of the behaviours are available as online supplemental material.}
\label{fig:behaviours_cl}
\end{figure}

The difference between the fitness scores of the solutions achieved with $\mathbf{b_{cm}}$ and $\mathbf{b_{cl}}$ was not significant. Nevertheless, we found that the different characterisations affected the evolved behaviours. Many of the best solutions evolved with $\mathbf{b_{cl}}$ were similar to the solutions evolved by fitness-based evolution, namely the formation and disbandment of small clusters. However, new behaviour patterns were also found. The distinctive behaviours evolved by novelty search with $\mathbf{b_{cl}}$ were focused on the exploitation of inter-robot relations and of flocking in particular. For example, the following two behaviour patterns were identified (see Figure~\ref{fig:behaviours_cl}):

\begin{description}
\item[\bf cl1:] The robots navigate in circles, and when two robots meet, one starts to follow the other, which leads to flocking with circular trajectories. Eventually, a single moving file is formed.
\item[\bf cl2:] Similar to (a), but when a robot cluster reaches a reasonable size, the cluster becomes static.
\end{description}

Overall, novelty search with $\mathbf{b_{cl}}$ focused on different classes of behaviours than novelty search with $\mathbf{b_{cm}}$. One of the main reasons for the difference in the evolved behaviours is conflation. Conflation occurs when individuals with distinct observable behaviours have very similar behaviour characterisation vectors~\citep{stan11a}. The consequence is that an individual with a distinct observable behaviour might not be considered novel by the novelty measure, and may thus disappear from the population. Conflation can represent both an advantage in terms of efficiency because it reduces the size of the search space, and a disadvantage when it inhibits the discovery of successful solutions or important stepping stones.

\begin{figure}[h]
\centering 
	\includegraphics[width=0.8\textwidth]{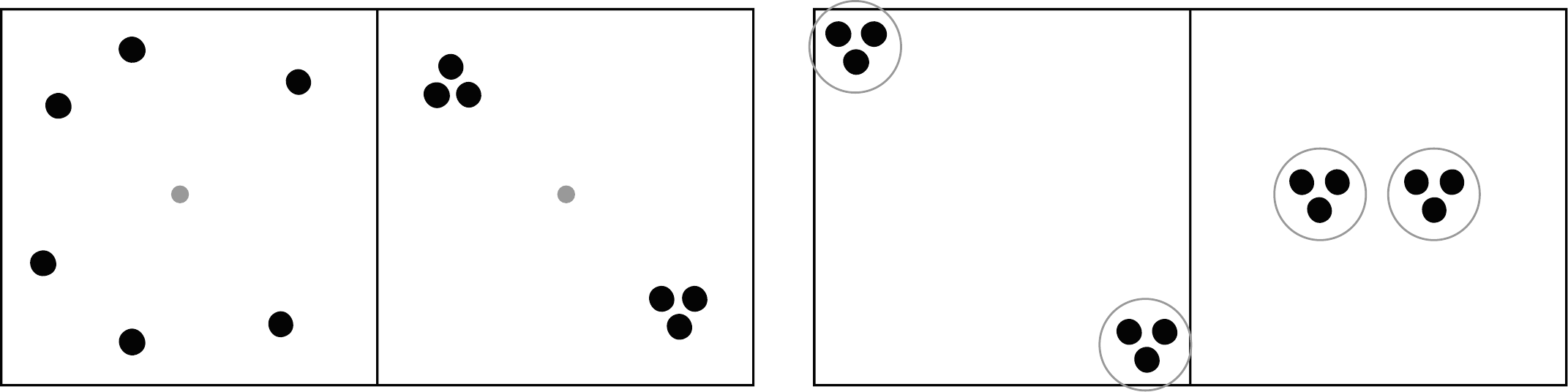}
\caption{An illustration of conflation for the centre of mass behaviour characterisation $\mathbf{b_{cm}}$ (left) and for the number of clusters behaviour characterisation $\mathbf{b_{cl}}$ (right). In both cases, the differences would not be reflected in the respective behaviour characterisations.}
\label{fig:conflation}
\end{figure}

Two examples of behaviours that can be conflated are shown in Figure~\ref{fig:conflation}. For the $\mathbf{b_{cm}}$ characterisation, the degree of clustering of the robots is irrelevant; while for the $\mathbf{b_{cl}}$ characterisation, the distance between robots (and clusters) is irrelevant. The impact of conflation can be seen in the evolved behaviours: with the $\mathbf{b_{cm}}$ characterisation, there were more behaviours that exploited walls, because navigating near them has a great impact on the novelty measure; while with $\mathbf{b_{cl}}$, the behaviours focused on the interactions between the robots and clusters, including following one another and disbanding clusters.

\subsubsection{Combined behaviour characterisation}
\label{sec:combined_aggregation}
Novelty search with the composed behaviour characterisation $\mathbf{b_{cmcl}}$ achieved on average marginally higher fitness scores than $\mathbf{b_{cm}}$ and $\mathbf{b_{cl}}$ (Mann--Whitney U test, $p$-value $<$ 0.05), see Figure~\ref{fig:ccm_graph}(right). The composed behaviour characterisation considers both the average distance to centre of mass and the number of clusters formed. As such, different behaviours are less likely to be conflated, which can potentially lead to better solutions.

It should, however, be noted that $\mathbf{b_{cm}}$ and $\mathbf{b_{cl}}$ are closely related with one another. It is not possible, for instance, to have a low average distance to the centre of mass and at the same time a large number of clusters. A large number of clusters implies that the swarm is scattered, and as such, the robots will have a high average distance to the centre of mass. Combining $\mathbf{b_{cm}}$ and $\mathbf{b_{cl}}$ reduces conflation to some extent, but since the two characterisations are related to one another, no additional effort is needed to explore the larger behaviour space. This explains why the behaviour space with more dimensions ($\mathbf{b_{cmcl}}$) did not have a negative impact in the effectiveness of novelty search in this case.

\subsection{Neural Network Complexity}

Previous work has shown that an advantage of novelty search, when used together with NEAT, is its ability to evolve solutions with lower genomic complexity \citep{stan11a}. To determine if such advantage holds in the aggregation task, we analysed the complexity (sum of the number of neurons and number of connections) of the solutions found by fitness-based evolution and by novelty search. The comparison is established by analysing the average complexity (across the multiple evolutionary runs) of the least complex solution with a fitness score above a certain threshold. The results are shown in Table~\ref{tab:complexity}. We only show the results for the $\mathbf{b_{cmcl}}$ characterisation because the results for the $\mathbf{b_{cm}}$ and $\mathbf{b_{cl}}$ characterisations are similar.

\begin{table}[b]
\caption{Comparison of the least complex solutions evolved by fitness-based evolution (Fit) and novelty search with $\mathbf{b_{cmcl}}$ (NS). The \emph{Complexity} columns lists the network complexity (sum of the number of neurons and the number of connections) of the least complex individual with a fitness score above a certain \emph{Fitness level}. The \emph{Generation} column lists the average generation in which the least complex individual was generated. The values are averages of 30 evolutionary runs for each method.}
\centering
\begin{tabular}{r r r r r r r}
\toprule
\multicolumn{ 1}{c}{} & \multicolumn{ 2}{c}{Generation} & \multicolumn{ 2}{c}{Complexity} \\ \cmidrule{ 2- 5}
\multicolumn{ 1}{r}{Fitness level} & \multicolumn{1}{r}{Fit} & \multicolumn{1}{r}{NS} & \multicolumn{1}{r}{Fit} & \multicolumn{1}{r}{NS} \\ \midrule
0.60 &  9 & 5  & 74.20 & 71.33 \\
0.65 &  18 & 7  & 75.77 & 71.67 \\
0.70 &  28 & 14 & 77.10 & 72.40 \\
0.75 &  42 & 22 & 79.10 & 74.13 \\
0.80 &  55 & 41 & 80.90 & 77.33 \\
0.85 &  76 & 63 & 83.17 & 81.70 \\
\bottomrule
\end{tabular}
\label{tab:complexity}
\end{table}

The results show that on average, for the same fitness levels, novelty search finds individuals with significantly less complex neural networks (Mann--Whitney U test, $p$-value $<$ 0.05). The difference is especially pronounced in the lower fitness levels. Note that the networks in the initial populations (without any hidden neurons) have a complexity of 71 (20 neurons and 51 links). The difference in the network complexity can be ascribed to the convergent nature of fitness-based evolution. If the best controllers in the earlier stages of evolution have more complex neural networks, fitness-based evolution starts to converge to such complex structures. As novelty search does not converge, and has tendency to explore simple solutions before moving on to more complex ones \citep{stan11a}, it is capable of finding solutions with a lower network complexity.

\section{Evolution of Resource Sharing Behaviours with Novelty Search}
\label{sec:sharing}

In this section, we study the application of novelty search to a more complex task in which a swarm of robots share a single resource. The swarm must coordinate in order to allow each member periodical access to a single battery charging station. In the task, the robots should first find the charging station, and then effectively share the station to ensure the survival of all the robots in the swarm. The charging station can only hold one robot at the time.

The problem of autonomous charging and resource conflict management is widely studied in the literature. \citet{cao97} identify resource conflicts as one of the fundamental challenges in the design of cooperative behaviours in multi-robot systems. Resource conflicts arise when a single indivisible resource (in our task, the charging station) is requested by multiple robots at the same time. The problem of sharing an energy charging station in particular is addressed in \citep{munoz02,robichaud02,kernbach11}.

\subsection{Experimental Setup}
\label{sec:energy_setup}

\begin{SCfigure}[1.3][b]
\centering 
	\includegraphics[width=0.35\textwidth]{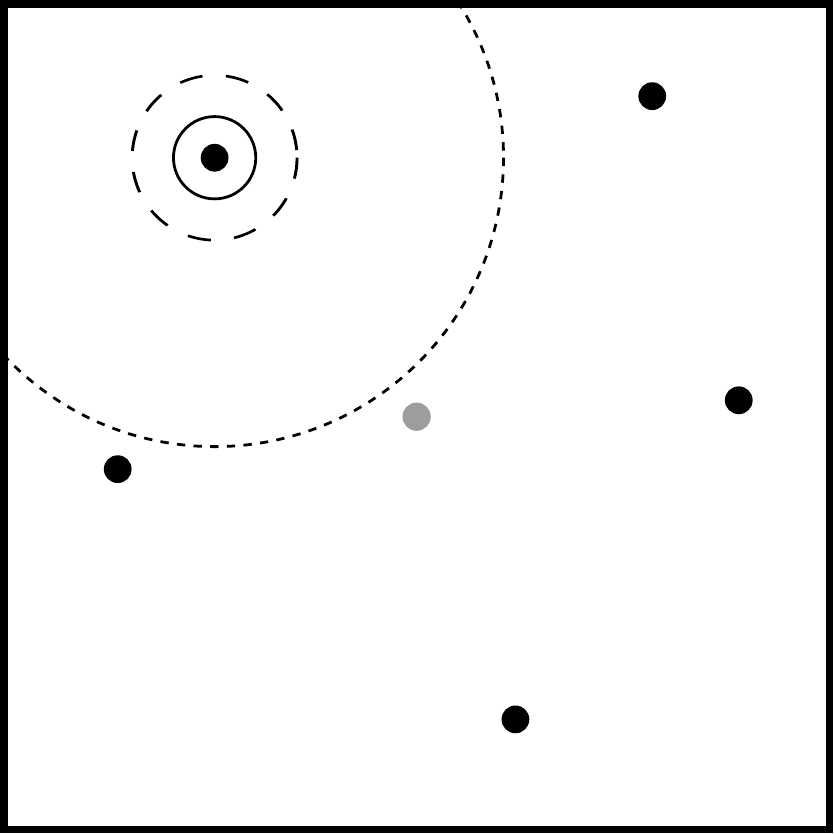}
\caption{The resource sharing task experimental setup. The grey circle in the centre is the charging station. The black filled circles are the robots (starting positions vary in each simulation). The solid circle around the top left robot represents the range of the obstacles sensor, the dashed circle represents the range of the robot sensor, and the fine dashed circle represents the range of the charging station sensor.}
\label{fig:energy_setup}
\end{SCfigure}

We use a swarm composed of 5 homogeneous robots. The robots are identical to the ones used in the aggregation experiments (see Section~\ref{sec:aggregation_setup}), except that they are equipped with additional sensors. Each robot has (i)~8 IR sensors evenly distributed around its chassis for the detection of obstacles (walls or other robots) up to a range of 10\,cm; (ii)~8 sensors dedicated to the detection of other robots up to a range of 25\,cm; (iii)~a ring of 8 sensors for the detection of the charging station up to a range of 1\,m; (iv)~a binary sensor that indicates whether or not the robot is currently being recharged; and (v)~a proprioceptive sensor that reads the current energy level of the robot. The sensors (i), (ii) and (iii) return the distance of the object that is being sensed. The experimental setup and the sensor ranges are depicted on Figure~\ref{fig:energy_setup}. The environment is a 3\,m by 3\,m square arena bounded by walls.

Each robot starts with full energy (1000 units), and the energy consumption increases linearly with the speed of the motors: a robot spends 5 units per second when motors are off, and 10 units of energy per second when both motors operate at their maximum speed. The charging station has the same diameter as a robot, and to recharge, the robots must remain static inside the charging station. The charging station is located in the centre of the arena, and charges a robot at a rate of 100 units of energy per second. Similarly to the aggregation experiments, each controller is evaluated in 10 simulations with random starting positions for the robots. Each simulation lasts for 2500 simulation steps. The highest scoring individual of each generation was post-evaluated in 100 simulations.

\subsection{Configuration of the Evolutionary Algorithms}
\label{sec:energy_functions}

We used the same parameter values for novelty search and the NEAT algorithm as in the aggregation experiments (see Section~\ref{sec:aggregation_algsetup}). However, we continued each evolutionary run until the 400th generation, since the resource sharing task proved to be more challenging than the aggregation task.

The fitness function $F_{s}$ used to evaluate the controllers is a linear combination of the number of robots alive at the end of the simulation (henceforth referred to as \emph{survivors}) and the average energy of the robots throughout the entire simulation:
\begin{equation}
F_{s} = 0.9\cdot\frac{|a_{T}|}{N}+0.1\cdot \sum_{t=1}^{T}\sum_{i=1}^{N}\frac{e_{i,t}}{TNe_{max}}\enspace ,
\end{equation}
where $|a_{T}|$ is the number of survivors, $T$ is the length of the simulation, $N$ is the number of robots in the swarm, $e_{i,t}$ is the energy of the robot $i$ at time $t$, and $e_{max}$ is the energy capacity of a robot. The second term of $F_s$ concerning the average energy was included to differentiate solutions where the same number of robots survive. Without the second term, there would be no fitness gradient, since it is unlikely that the initial population contains solutions where at least one robot survives until the end.\footnote{We empirically determined the probability of randomly generating a solution where at least one robot survives until the end (in any of the 10 trials) to be approximately 1\%.}

We experimented with behaviour characterisations of a different nature in the resource sharing experiments, compared to the characterisations used in the aggregation experiments. While in the aggregation task, we used spatial relationships between the robots sampled every 5\,s during the simulation, in the resource sharing experiments, we use only quantities that characterise a simulation as a whole. We evaluated two behaviour characterisations:

\begin{description}
\item[$\mathbf{b_{simple}}$:] The first characterisation is closely related to the fitness function, and it is composed of two values (normalised to the interval $[0,1]$): (i)~the number of robots that survive till the end of the simulation; and (ii)~the average energy of all alive robots throughout the simulation. $\mathbf{b_{simple}}$ is given by:
\begin{equation} \label{eq:bsimple_energy}
\mathbf{b_{simple}} = \left ( \frac{|a_{T}|}{N} \;,\; \sum_{t=1}^{A} \sum_{i \in a_{t}}^{ }\frac{e_{i,t}}{A\cdot |a_{t}| \cdot e_{max}} \right ) \enspace ,
\end{equation}
where $A$ is the number of time steps in which there was at least one robot alive and $a_{t}$ is the set of robots alive at time $t$.

\item[$\mathbf{b_{extra}}$:] The second behaviour characterisation is an extension of $\mathbf{b_{simple}}$. Two more features, which are not directly related to the fitness function, were added to the characterisation: (i)~the average speed of the alive robots throughout the simulation; and (ii)~the average distance of the alive robots to the charging station. The movement of a robot in a given instant is determined by the average wheel speed at that instant. The two additional features are also normalised to the interval $[0,1]$. $\mathbf{b_{extra}}$ is defined as:
\begin{equation} \label{eq:bextra_energy}
\mathbf{b_{extra}} = \left (\mathbf{b_{simple}} \;,\; \sum_{t=1}^{A} \sum_{i \in a_{t}}^{ }\frac{s_{i,t}}{A\cdot |a_{t}| \cdot s_{max}} \;,\; \sum_{t=1}^{A} \sum_{i \in a_{t}}^{ }\frac{d_{i,t}}{A\cdot |a_{t}| \cdot d_{max}} \right ) \enspace ,
\end{equation}
where $s_{i,t}$ and $d_{i,t}$ are the speed of the robot $i$ and its distance to the charging station, respectively, at time $t$. $s_{max}$ is the maximum speed of a robot, and $d_{max}$ is half the length of the diagonal of the arena.
\end{description}

\subsection{Performance Comparison}

\begin{figure}[b]
	\centering
	\includegraphics[width=1\textwidth]{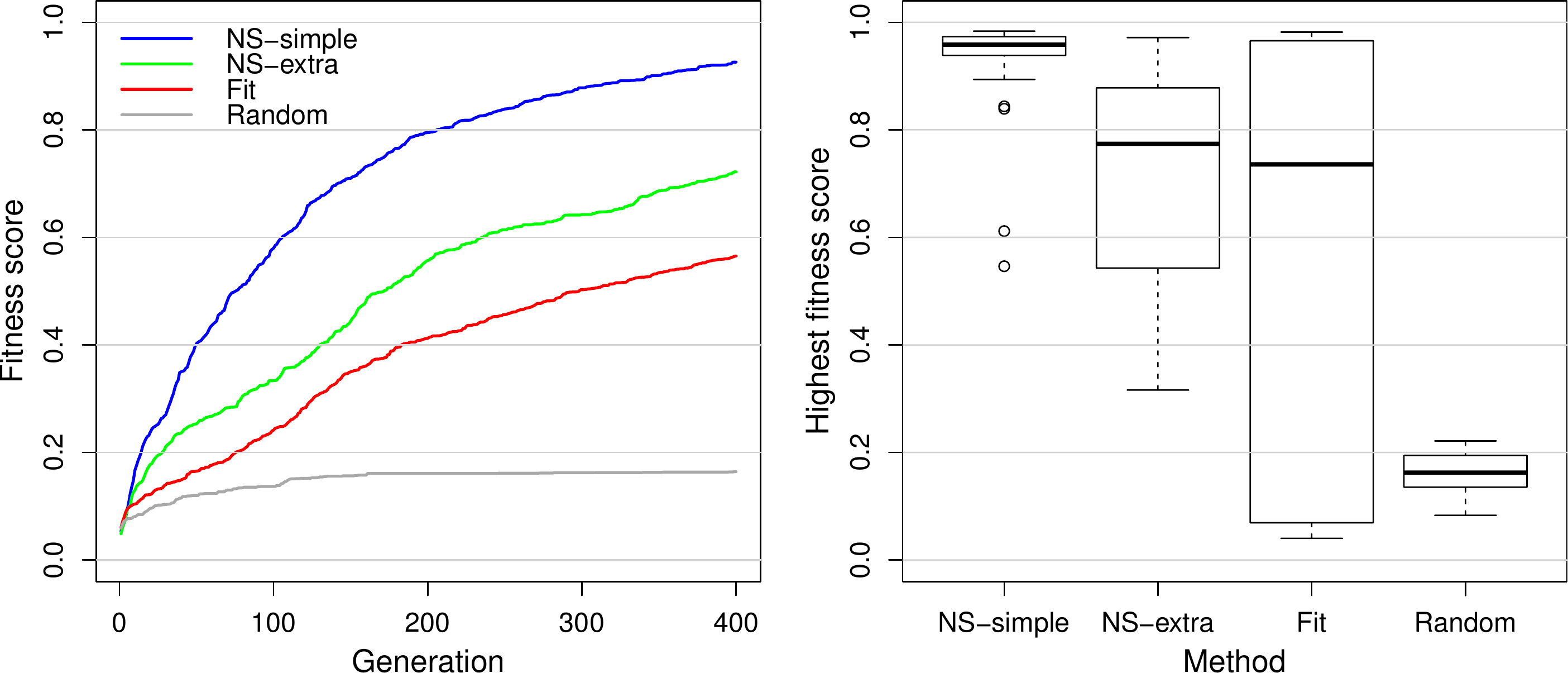}
\caption{Highest fitness scores achieved in the resource sharing task with novelty search with $\mathbf{b_{simple}}$ and $\mathbf{b_{extra}}$ (\emph{NS-simple}, \emph{NS-extra}), fitness-based evolution (\emph{Fit}), and random evolution (\emph{Random}). Left: average fitness value of the highest scoring individual found so far at each generation. The values are averaged over 30 independent evolutionary runs for each method. Right: box-plots of the highest fitness score found in each evolutionary run, for each method. The whiskers extend to the lowest and the highest data point within 1.5 times the interquartile range. Outliers are indicated by circles. The maximum fitness score in practice is about 0.98, which corresponds to all robots surviving until the end of the experiment, while maintaining high levels of energy.}
\label{fig:energy_fitness}
\end{figure}

Figure~\ref{fig:energy_fitness} depicts the highest fitness scores achieved by novelty search, fitness-based evolution, and random evolution. Random evolution only achieves very low fitness scores ($<0.2$). In fitness-based evolution, the distribution of the highest fitness scores achieved is characteristically wide. Fitness-based evolution achieved close to the maximum fitness score in 10/30 of the evolutionary runs, but failed to evolve any viable solution (where at least one robot consistently survives) in 10/30 of the runs. Bootstrapping proved difficult in fitness-based evolution, as most runs got stuck in low regions of the fitness landscape for a large number of generations.

Novelty search, on the other hand, did not get stuck in local maxima. The result indicates that novelty search was unaffected by deception, and was capable of bootstrapping the evolutionary process. Novelty search with $\mathbf{b_{simple}}$ consistently achieved high fitness scores, with 4--5 robots surviving in the best solutions. The fitness scores achieved with $\mathbf{b_{simple}}$ are significantly higher than those achieved by fitness-based evolution (Mann--Whitney U test, $p$-value $<$ 0.01). With $\mathbf{b_{extra}}$, novelty search failed to match the performance of $\mathbf{b_{simple}}$, and the fitness scores achieved are significantly lower ($p$-value $<$ 0.01). Within 400 generations, \emph{NS-simple} could consistently achieve fitness scores close to the maximum value. However, it should be noted that \emph{NS-extra} and \emph{Fit} do not appear stable at that point, and if more generations were allowed, they might still improve. The differences between the behaviour characterisations and the evolved behaviours are discussed below.

\subsection{Behavioural Diversity}

\begin{figure}
\centering 
	\includegraphics[width=1\textwidth]{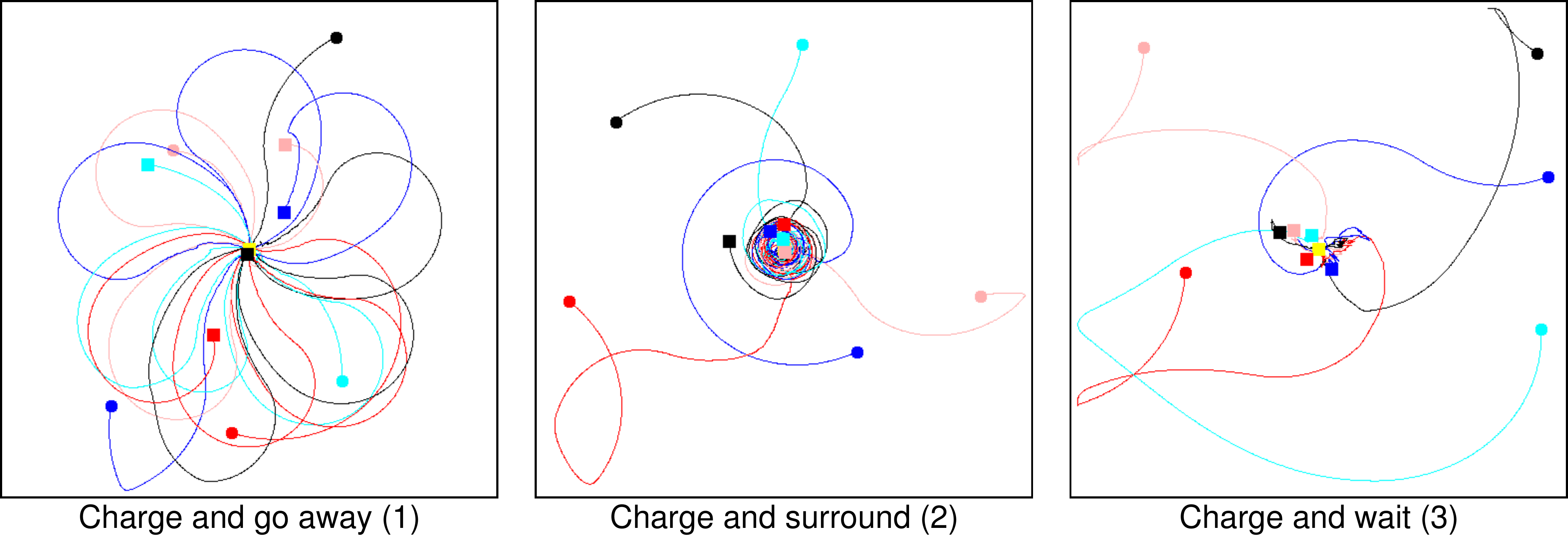}
\caption{The patterns of behaviour corresponding to the highest scoring solutions found by novelty search and fitness-based evolution in the resource sharing task. Each line represents the trajectory of the robot throughout the simulation. The circles indicate initial positions and squares indicate final positions. Videos of the behaviours are available as online supplemental material.}
\label{fig:behaviours_energy}
\end{figure}

Through visual inspection of the solutions that achieved the highest fitness scores (above 0.95), we found that all evolutionary methods produced solutions that display similar behaviours. In the successful behaviours, the robots always start by searching for the charging station. Depending on the solution, the robots move in straight lines, in large circles, or in spirals, until the station has been located. The second part of the successful behaviours concerns the coordination of access to the charging station. We observed three different coordination behaviours (see Figure~\ref{fig:behaviours_energy}):

\begin{description}
\item[\textbf{Charge and go away:}] If a robot in the charging station detects another robot approaching, it leaves the station to let the approaching robot recharge. The leaving robot performs what resembles a random walk in the arena and eventually returns to the station to recharge.
\item[\textbf{Charge and surround:}] If a robot in the charging station detects another robot approaching, it leaves the station to let the approaching robot recharge.  The leaving robot begins to circle the charging station, and approaches the station again when its energy level is below a solution-specific threshold.
\item[\textbf{Charge and wait:}] Once a robot is in the charging station, it continues to occupy the station until its energy level is above a solution-specific threshold. When the robot leaves, it moves only a short distance away from the station. Then, the robot remains almost static until its energy level is below a solution-specific threshold, at which point the robot tries to recharge again.
\end{description}

\subsubsection{Fitness-based evolution}

As mentioned above, fitness-based evolution did not consistently evolve solutions with high fitness scores (see Figure~\ref{fig:energy_fitness}). In fact, of the 30 runs conducted, 10 runs never evolved solutions with a fitness score much higher than the initial randomly generated population. We analysed the controllers evolved in the runs that only achieved low fitness scores and identified two behaviour patterns:
\begin{itemize}
\item The robots move slowly in small circles. When one of the robots detects the charging station, it moves towards the station, and occupies it till the end of the simulation. As a result, the rest of the swarm dies.
\item All the robots remain almost static from the beginning of the simulation until they run out of energy.
\end{itemize}
Both behaviours represent local maxima in the fitness landscape. In the first case, evolution converges to solutions where only one robot survives at the expense of the rest of the swarm. In the second case, the evolutionary process starts to converge to controllers that reduce the wheel speed to conserve energy. Conserving energy causes the robots to survive longer, thus slightly increasing the fitness score of the controller. However, reducing the wheel speed also decreases the chance that a robot will find the charging station. Once an evolutionary process starts to converge to a local maximum based on energy conservation, it can take many generations to escape from that maximum, or evolution may not escape at all.

\subsubsection{Simple behaviour characterisation}

\begin{figure}
\centering
\includegraphics[width=\textwidth]{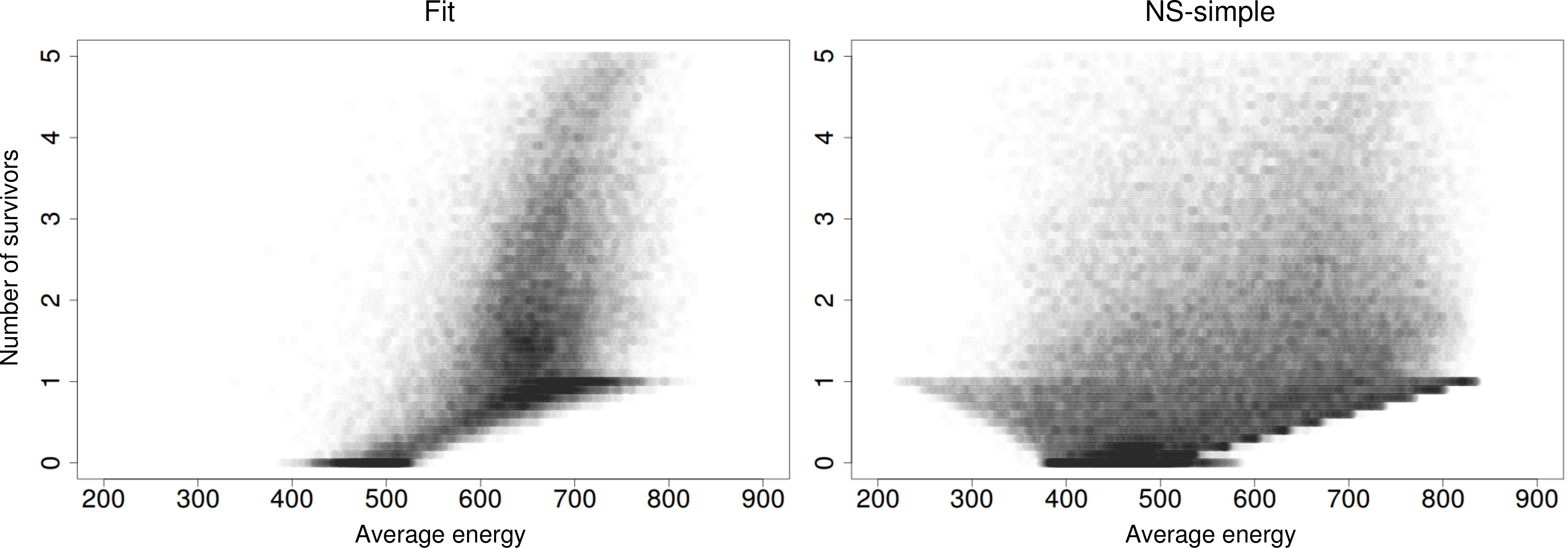}
\caption{Behaviour space exploration with fitness-based evolution (\emph{Fit}) and novelty search with $\mathbf{b_{simple}}$ (\emph{NS-simple}), in all evolutionary runs. The $x$-axis is the average energy level of the robots still alive, the $y$-axis is the number of survivors. Each individual is mapped according to its characterisation. Darker zones indicate that there were more individuals evolved with the behaviour of that zone.}
\label{fig:space_simple}
\end{figure}

The highest scoring solutions evolved by novelty search follow the same behaviour patterns as the highest scoring solutions evolved by fitness-based evolution. However, an analysis of the explored behaviour space reveals that novelty search could in fact evolve a greater diversity of solutions (see Figure~\ref{fig:space_simple}). The greater diversity comes from variations of the same behaviour patterns. For instance, if the coordination behaviours use different energy thresholds to trigger entering and leaving the charging station, it will result in different average levels of energy. In such cases, the behaviour patterns may appear similar through visual inspection, but they have distinct behaviour characterisations.

\subsubsection{Extra behaviour characterisation}

\begin{SCfigure}[1.2]
\centering
\includegraphics[width=0.5\textwidth]{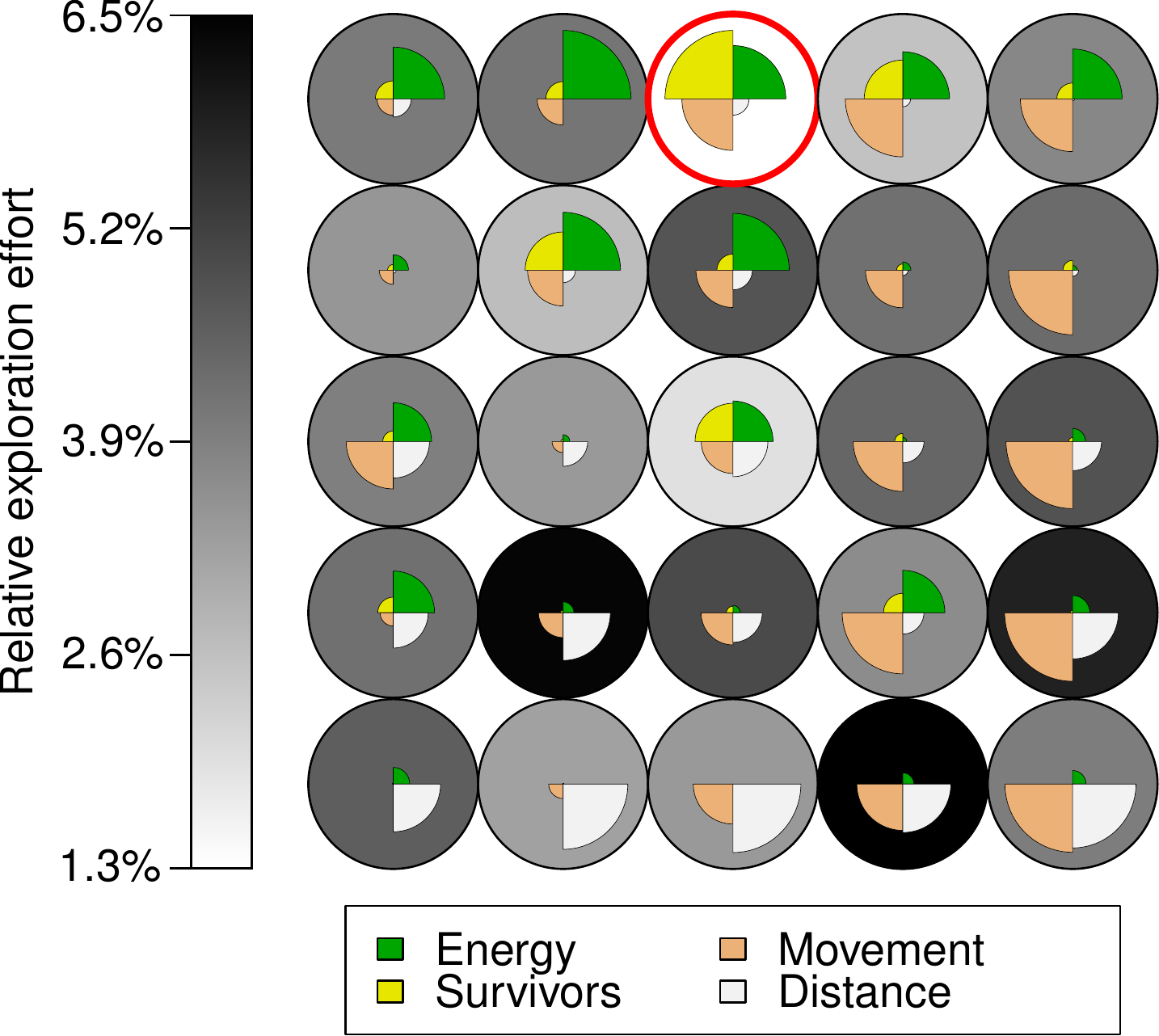}
\caption{Kohonen map representing the explored behaviour space in novelty search with $\mathbf{b_{extra}}$. Each circle represents a behaviour pattern, depicted by the 4 slices of different colour. Each slice represents one component of the behaviour characterisation vector -- the bigger the slice, the higher the value of that component. The darker the background of a circle, the more individuals were evolved with the corresponding behaviour.}
\label{fig:extra_som}
\end{SCfigure}

We adapted the visualisation technique based on Kohonen maps to the 4-dimensional behaviour space created by the $\mathbf{b_{extra}}$ characterisation to analyse the degree of exploration of different behaviour regions. The results in Figure~\ref{fig:extra_som} show how novelty search with the $\mathbf{b_{extra}}$ characterisation explored the behaviour space --- a notable variety of combinations of average energy, movement, and distance to the charging station. However, the behaviour dimension related to the number of surviving robots was the least explored dimension of the behaviour space. In almost all the explored regions, the number of surviving robots was either zero or one. Only a single behaviour region included behaviours in which the whole swarm often survives (highlighted in Figure~\ref{fig:extra_som}), and that region was one of the least explored.

The relatively low performance of novelty search is caused by the significantly larger behaviour space and the lack of correlation between the behaviour features. Novelty search can freely explore dimensions such as average movement and distance to charging station without any robots surviving. Furthermore, both of these dimensions are intuitively easier to explore than the dimension concerning the number of surviving robots. As such, the opportunistic nature of evolution causes the search to focus less on the surviving robots dimension, often inhibiting the evolution from finding successful solutions to the task. This is in contrast with the correlated features used in the $\mathbf{b_{cmcl}}$ characterisation from the aggregation experiments, in which novelty search performed well despite the enlarged behaviour space (see Section~\ref{sec:combined_aggregation}).

The two-dimensional $\mathbf{b_{simple}}$ characterisation restricts novelty search to explore the two dimensions directly related to the fitness function. Similar solutions in which robots move at different speeds are, for instance, conflated when the $\mathbf{b_{simple}}$ characterisation is used, whereas they are considered different (and potentially novel) when the $\mathbf{b_{extra}}$ characterisation is used. As a consequence, novelty search with the $\mathbf{b_{simple}}$ characterisation focused exploration on the dimensions directly related to the fitness function and had significantly more success in finding solutions with high fitness scores. On the other hand, conflation can be prejudicial to the diversity of solutions that is evolved. For instance, analysing the best solutions evolved with each characterisation, we verified that the behaviour pattern  \emph{Charge and wait} was less common with $\mathbf{b_{simple}}$ when compared to $\mathbf{b_{extra}}$. This behaviour is intuitively one of the most interesting solutions, since it takes advantage of the variable energy spending to extend the life time of the robots. However, it was conflated with $\mathbf{b_{simple}}$, since this characterisation does not consider the speed of the robots.

Our results suggest that caution must be displayed when behaviour characterisations are defined. Dimensions that novelty search may opportunistically explore at the cost of other dimensions that are more crucial for the task should not be included. Alternatively, a number of methods have been proposed that aim to guide exploration in novelty search towards high fitness solutions. In Section~\ref{sec:combination}, we study the application of two such methods to the resource sharing task.

\subsection{Neural Network Complexity}

The experiments with the aggregation task showed that novelty search found successful solutions with less complex neural networks than fitness-based evolution. To determine if the same is true for the resource sharing task, we analysed the complexity of the solutions evolved for this task. Table~\ref{tab:complexity_rs} shows a comparison between the complexity of the solutions found by novelty search with $\mathbf{b_{simple}}$ and fitness-based evolution. The results show that there was a considerable difference between the complexity of the solutions evolved by the two methods. At fitness levels above 0.2, the simpler solutions evolved by novelty search were of significantly lower complexity than the solutions evolved by fitness-based evolution (Mann--Whitney U test, $p$-value $<$ 0.05). Note that the complexity of the networks in the initial populations is 107 (29 neurons and 78 links).

\begin{table}
\caption{Comparison of the least complex solutions evolved by fitness-based evolution (Fit) and novelty search with $\mathbf{b_{simple}}$ (NS). The \emph{Complexity} columns lists the network complexity (sum of number of neurons and number of connections) of the least complex individual with a fitness score above a certain \emph{Fitness level}. The \emph{Generation} column lists the average generation in which the least complex individual was generated. The values are averages of 30 evolutionary runs for each method.}
\centering
\begin{tabular}{r r r r r r r}
\toprule
\multicolumn{ 1}{c}{} & \multicolumn{ 2}{c}{Generation} & \multicolumn{ 2}{c}{Complexity} \\ \cmidrule{ 2- 5}
\multicolumn{ 1}{r}{Fitness level} & \multicolumn{1}{r}{Fit} & \multicolumn{1}{r}{NS} & \multicolumn{1}{r}{Fit} & \multicolumn{1}{r}{NS} \\ \midrule
0.20 &  65 & 21  &  114.90 &  110.50 \\
0.40 &  150 & 69  & 127.30 & 118.40 \\
0.60 &  192 & 135 &  134.20 & 125.90 \\
0.80 & 213 & 192 & 136.90 & 132.40 \\
0.90 &  268 & 239 & 145.70 & 133.90 \\
\bottomrule
\end{tabular}
\label{tab:complexity_rs}
\end{table}

\section{Combining Novelty and Fitness}
\label{sec:combination}

The results of the resource sharing experiments showed that although novelty search finds a broad diversity of behaviours, regions with high-fitness behaviours may never be explored. This issue has also been reported in other studies \citep{stan10a,cuccu11b}, and a common solution is to combine novelty search with fitness-based evolution (see Section~\ref{sec:ns_variants}). The combination is promising, because while novelty search promotes exploration of the behaviour space, fitness-based evolution exploits existing high fitness solutions. In this section, we study the application of \emph{progressive minimal criteria novelty search} (PMCNS) \citep{gomes12} and \emph{linear scalarization} \citep{cuccu11b} to the evolution of solutions for the resource sharing task. We compare the results obtained with PMCNS and with linear scalarization to the results obtained with pure novelty search and with fitness-based evolution.

\subsection{Progressive Minimal Criteria Novelty Search}

PMCNS~\citep{gomes12} is an extension of \emph{minimal criteria novelty search} (MCNS)~\citep{stan10a}. In MCNS, the exploration of the behaviour space is restricted by domain-dependent minimal criteria that evolved individuals must meet in order to be selected for reproduction. The objective of PMCNS is to take advantage of the restrictions on behaviour space exploration provided by MCNS, but without having to define fixed, domain-dependent criteria \emph{a priori}. In PMCNS, a dynamic fitness threshold is used as the minimal criterion: individuals with a fitness score above the threshold meet the criterion. The fitness threshold is progressively increased during the evolutionary process. The idea behind the increasing fitness criterion is to progressively restrict the search space to regions of the behaviour space with higher fitness scores, and thereby avoid that novelty search spends much or all effort on regions of the behaviour space with novel, but low fitness behaviours.

The minimal criterion starts at the theoretical minimum of the fitness score (typically zero), so all controllers initially meet the criterion. In each generation $g$, the new criterion $mc_{g}$ is based on the fitness score $v_{g}$ of the $P$-th percentile of the individuals in the current population, i.e., the fitness score below which $P$ percent of the individuals fall. The parameter $P$ controls the exigency of the minimal criterion (0 -- all individuals meet the criterion, 1 -- only the individual with the highest fitness meets the criterion). Only increases in the minimal criterion are allowed, and in order to smoothen the increase, the minimal criterion from the previous generation is used to determine the criterion for the current generation (Eq.~\ref{eq:pmcns1}). The score used for selection of each individual $i$ in the population is then calculated according to Eq.~\ref{eq:pmcns2}.

\begin{equation}
mc_{g} = mc_{g-1} + \mathrm{max}(0, (v_{g} - mc_{g-1}) \cdot S)
\label{eq:pmcns1}
\end{equation}

\begin{equation}
\mathrm{score}(i) =	
\begin{cases}
	nov_{i} & \text{if $f\!it_{i} \geq mc_{g}$}
	\\
	0 & \text{otherwise}
\end{cases}
\label{eq:pmcns2}
\end{equation}

The variables $nov_{i}$ and $f\!it_{i}$ are the novelty and fitness score of the individual $i$, respectively. The smoothening parameter $S$ controls the speed of the adaptation of the minimal criterion.


\subsection{Linear Scalarization of Novelty and Fitness Scores}

\citet{cuccu11b} proposed a linear scalarization of novelty and fitness score, as an approach to sustain diversity while improving the performance of traditional fitness-based evolution. Each individual $i$ is evaluated to obtain both fitness score, $f\!it(i)$, and novelty score, $nov(i)$, which after being normalised (Eq.~\ref{eq:blendnorm}) are combined according to Eq.~\ref{eq:blend}.

\begin{equation}
\overline{f\!it}(i)=\frac{f\!it(i)-f\!it_{min}}{f\!it_{max}-f\!it_{min}}, \quad \overline{nov}(i)=\frac{nov(i)-nov_{min}}{nov_{max}-nov_{min}} \enspace ,
\label{eq:blendnorm}
\end{equation}

\begin{equation}
score(i)=(1-\rho)\cdot \overline{f\!it}(i)+\rho \cdot \overline{nov}(i) \enspace .
\label{eq:blend}
\end{equation}

The parameter $\rho$ controls the relative weight of fitness and novelty, and must be specified by the experimenter (usually through trial and error). $f\!it_{min}$ and $nov_{min}$ are respectively the lowest fitness and novelty scores in the current population, and $f\!it_{max}$ and $nov_{max}$ are the corresponding highest scores.

\subsection{Experimental Setup}

We used the experimental setup of the resource sharing task (see Section~\ref{sec:energy_setup}) for this set of experiments. The fitness function and the behaviour characterisations are also the same as we used previous experiments (see Section~\ref{sec:energy_functions}).

For PMCNS, we chose a percentile value of $P=0.50$ (values of 0.25, 0.50 and 0.75 were tested), which corresponds the median value of the fitness scores in the population, and a smoothening parameter of $S=0.25$, which corresponds to relatively slow increases in the minimal criterion value. For linear scalarization, the parameter $\rho$ was set to $\rho=0.75$ (values of 0.25, 0.50 and 0.75 were tested), which means that the score of each individual is composed of 75\% of the novelty score and 25\% of the fitness score. A discussion of the parameter values for linear scalarization and PMCNS can be found in \citep{cuccu11b} and in \citep{gomes12}, respectively.

\subsection{Results}

\begin{figure}[b]
\centering
\includegraphics[width=\textwidth]{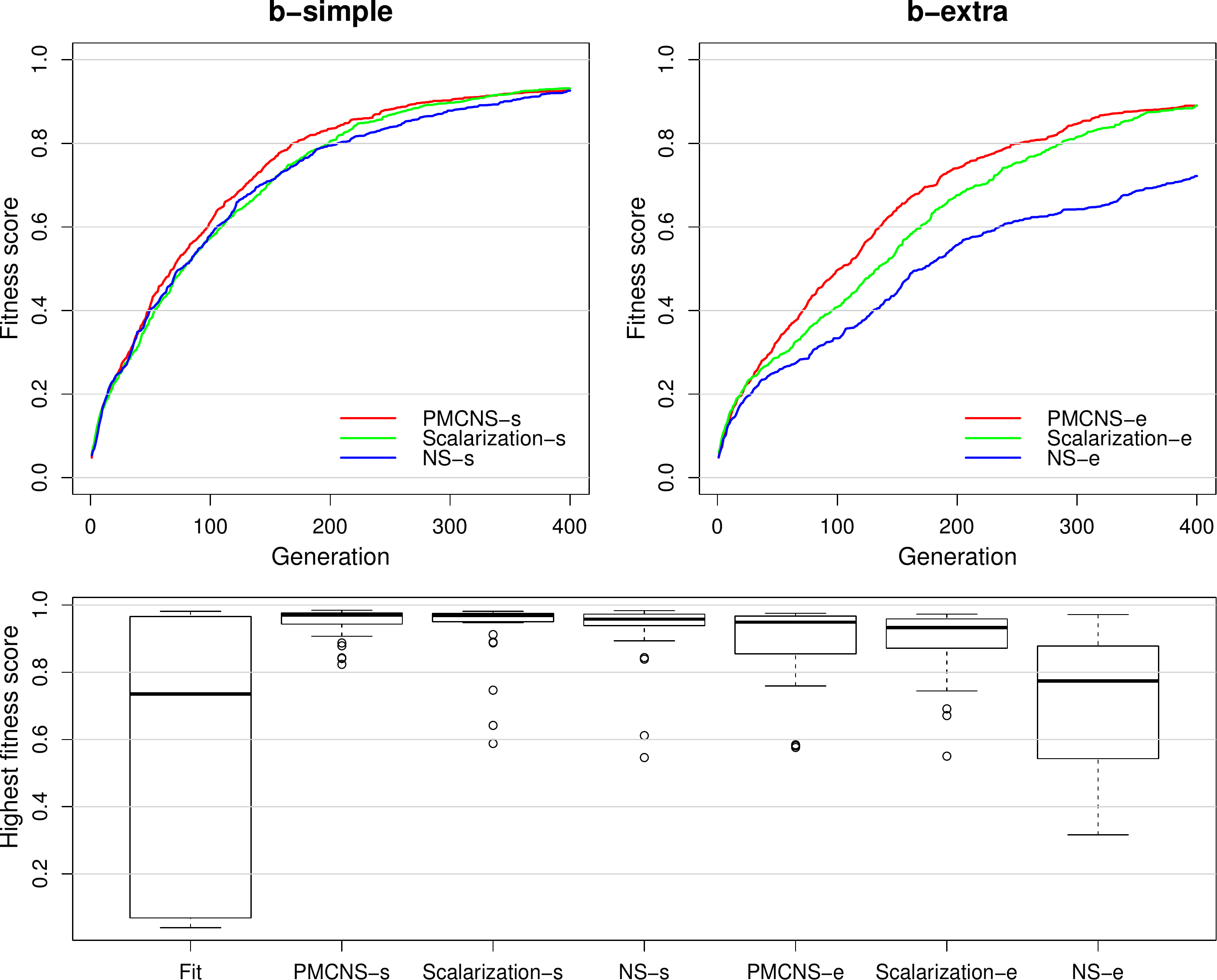}
\caption{Highest fitness scores obtained in the resource sharing task with fitness-based evolution (\emph{Fit}), and novelty-based evolutionary techniques with $\mathbf{b_{simple}}$ (\emph{PMCNS-s}, \emph{Scalarization-s}, \emph{NS-s}) and $\mathbf{b_{extra}}$ (\emph{PMCNS-e}, \emph{Scalarization-e}, \emph{NS-e}). Top: average fitness value of the highest scoring individual found so far at each generation. The values are averaged over 30 independent evolutionary runs for each method. Bottom: box-plots of the highest fitness score found in each evolutionary run, for each method. The whiskers extend to the lowest and the highest data point within 1.5 times the interquartile range. Outliers are indicated by circles.}
\label{fig:mix_fitness_trajectories}
\end{figure}

We ran experiments with both behaviour characterisations $\mathbf{b_{simple}}$ and $\mathbf{b_{extra}}$ to evaluate the performance of PMCNS and linear scalarization in behaviour spaces of different dimensionality. The resulting fitness trajectories are shown in Figure~\ref{fig:mix_fitness_trajectories}.

The results show that both PMCNS and linear scalarization are more effective than pure novelty search when using $\mathbf{b_{extra}}$ (Mann--Whitney U test, $p$-value~$<$~0.01), and achieve similar fitness scores when using $\mathbf{b_{simple}}$. With the $\mathbf{b_{extra}}$ characterisation, pure novelty search fails to reach high fitness scores, as evolution tends to focus the exploration on behaviour dimensions that are not directly relevant for solving the task. The results suggest that PMCNS and linear scalarization can overcome this issue, and that the inclusion of a fitness component in the selection criteria helps to guide the evolutionary process towards solutions with high fitness. It should also be noted that PMCNS and linear scalarization do not appear to be affected by the deceptiveness of the fitness function since they could consistently achieve high fitness scores.

\begin{figure}[b]
\centering
\includegraphics[width=\textwidth]{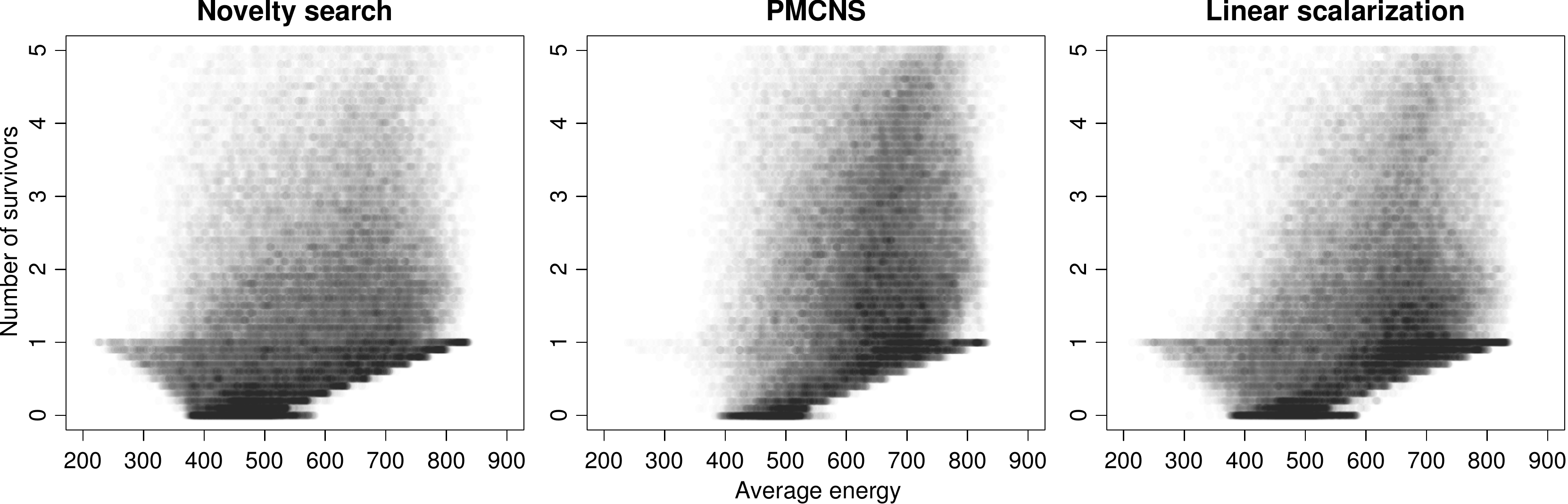}
\caption{Behaviour space exploration for each variant of novelty search, with the $\mathbf{b_{simple}}$ characterisation, in all evolutionary runs. The $x$-axis is the average energy level of the robots still alive, the $y$-axis is the number of survivors. Each individual is mapped according to its behaviour. Darker zones indicate that there were more individuals evolved with the behaviour of that zone.}
\label{fig:space_exploration_simple}
\end{figure}

\begin{figure}[b]
\centering
\includegraphics[width=\textwidth]{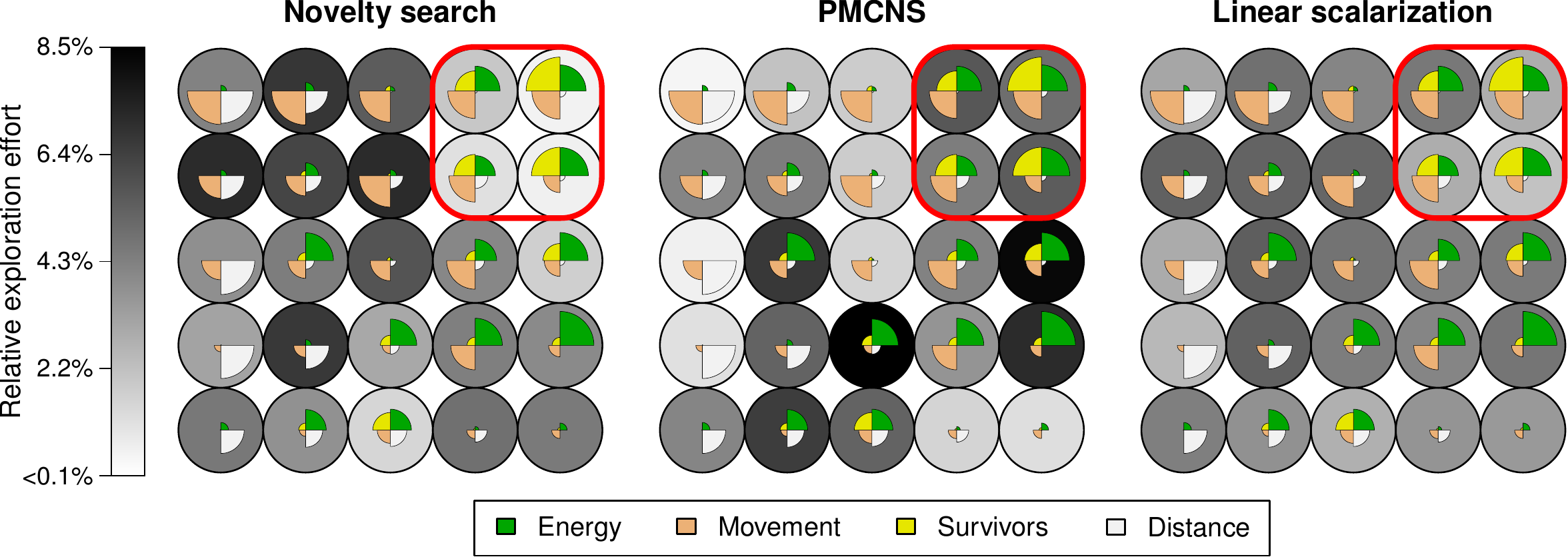}
\caption{Kohonen maps representing the explored behaviour space with each variant of novelty search, with the $\mathbf{b_{extra}}$ characterisation. Each circle represents a behaviour pattern, depicted by the 4 slices of different colour. Each slice represents one component of the behaviour characterisation -- the bigger the slice, the greater the value of that component. The darker the background of a circle is, the more individuals were evolved with the corresponding behaviour. The behaviour patterns with higher fitness scores are indicated in the upper right corner of each map.}
\label{fig:space_exploration_extra}
\end{figure}

Figure~\ref{fig:space_exploration_simple} and Figure~\ref{fig:space_exploration_extra} depict the behaviour space exploration with $\mathbf{b_{simple}}$ and $\mathbf{b_{extra}}$, respectively. An analysis of the behaviour space exploration shows that PMCNS and linear scalarization have a greater focus on regions associated with high fitness scores compared to pure novelty search. The coverage of the behaviour space was not negatively affected in PMCNS and linear scalarization. In fact, the two methods found the same broad range of behaviours as pure novelty search. The difference between pure novelty search and novelty search combined with fitness-based search is the amount of exploration done in each behaviour region: PMCNS and linear scalarization focused less on the low fitness behaviours (few or no robots surviving till the end of the simulation) and more on the high-fitness behaviours (high number of surviving robots). As a result, the PMCNS and linear scalarization could achieve solutions with significantly higher fitness scores than the best solutions evolved with pure novelty search.

\section{Conclusions}
\label{sec:conclusion}

We studied the application of novelty search to the evolution of controllers for swarms of robots. The study was based on two distinct swarm robotics tasks: (i)~an aggregation task, and (ii)~a resource sharing task. The aggregation task was non-deceptive, as fitness-based evolution consistently managed to find high fitness solutions. Nevertheless, novelty search could achieve a similar performance in terms of fitness scores. We also showed how novelty search found several alternative and successful solutions to the task. Our analysis was based on Kohonen self-organising maps that allowed for the visualisation of the degree of exploration conducted in different regions of the behaviour space. 

The resource sharing task was a deceptive setup in which fitness-based evolution often got stuck in local maxima. Novelty search was unaffected by deception and displayed a significantly better performance than fitness-based evolution. In both tasks, novelty search was distinctively able to bootstrap the evolutionary process, it could consistently find behaviours with high fitness scores early in the evolutionary process, and it was able to find successful solutions with lower neural network complexity than the solutions evolved by fitness-based evolution.

To the best of our knowledge, our study is the first in which novelty search has been applied to evolutionary swarm robotics. Since behaviour characterisations are domain-dependent and a fundamental component in novelty search, we studied two different approaches to the design of characterisations: one based on the spatial inter-robot relationships sampled at regular intervals, and one based on two to four quantities that summarise the swarm behaviour throughout an entire experiment. None of the characterisations depends on the swarm size and they are thus scalable. In our experiments, we combined different behaviour characterisations and found that such combinations were only effective when the dimensions in the characterisation were directly related to the task. The opportunistic nature of artificial evolution will cause the search to first focus on the behaviour dimensions that are easier to explore. If such dimensions are not related with the task, the search will spend considerable effort in unfruitful regions of the behaviour space, reducing the effectiveness of novelty search.

We discovered that novelty search may not always find high-fitness solutions, especially when the behaviour space has dimensions that are unrelated or only weakly related to the objective. To overcome this issue, we studied two variants of novelty search, PMCNS and linear scalarization of novelty and fitness scores, in which novelty search operates in concert with fitness-based evolution. Our experimental results showed that PMCNS and linear scalarization are effective in guiding the exploration towards behavioural regions of higher fitness, without compromising the capacity of novelty search to find a broad diversity of solutions. The diversity of evolved solutions was not negatively affected in PMCNS and linear scalarization, when compared to pure novelty search.

Overall, our study shows that novelty search can be successfully applied to swarm robotics systems. Novelty search has several advantages over fitness-based evolution. One of the prominent advantages is that novelty search often produces a broad diversity of successful behaviours based on self-organisation. In the swarm robotics domain, diversity and self-organisation are particularly important because of the difficulties of designing such behaviours by hand. Moreover, novelty search is unaffected by deception, less prone to bootstrapping issues, and can evolve solutions with less complex neural networks. Novelty search is therefore a promising alternative for the artificial evolution of controllers for swarm robotics systems. 

\subsection{Future work}

Our results showed that care must be taken in the definition of macroscopic behaviour characterisations, especially when combining distinct behaviour features. It is necessary to ensure that (i)~each component of the characterisation is relevant to the task, and (ii)~that the components do not differ too much in terms of how easy or hard it is for novelty search to explore the corresponding dimensions of the behaviour space. It may be difficult to guarantee that macroscopic behaviour characterisations always meet these criteria, and in ongoing work we are therefore studying how individual weights can be assigned to components and modified during evolution, either manually or automatically. Such weights could ensure that particularly important dimensions are thoroughly explored.

While pure novelty search performed reasonably well in our tasks, our results show that the inclusion of a fitness component can further increase the performance of novelty search. As such, in future work we are going to elaborate on strategies for combining novelty and fitness. For instance, we are studying if the use of dynamic weights in linear scalarization can bring significant advantages. Such approach could allow the search to focus more on exploitation or more on exploration at different stages of the evolutionary process.

An alternative that we are currently studying is \emph{generic novelty measures} for swarm robotics. \citet{doncieux10} proposed a behaviour characterisation based on the mapping between the sensor input and the actuator output. We are studying how such characterisations could be defined in the collective robotics domain, and if they represent a viable alternative to domain-dependent novelty measures. Such generic measures are typically used in combination with fitness-based evolution, since the behaviour spaces created by generic measures tend to be vast and only weakly related with the objective.

\section*{Acknowledgements}

This research has been supported by Funda\c{c}\~{a}o para a Ci\^{e}ncia e a Tecnologia (FCT) grants PTDC/EEACRO/104658/2008, PEst-OE/EEI/LA0008/2011, and SFRH/BD/89095/2012. 

\section*{Supplemental Material}

Videos of the behaviours are available as online supplemental material and at the following URL: \url{http://home.iscte-iul.pt/~alcen/si2013}. The source code of the software used in the experiments can be found at: \url{https://github.com/jorgemcgomes/evosimbad}.

\bibliographystyle{abbrvnat}
\bibliography{journal}

\end{document}